\RequirePackage{fix-cm}
\documentclass[twocolumn]{svjour3}          %
\smartqed  %
\usepackage{graphicx}
\usepackage{amsmath,amssymb} %

\usepackage{xcolor, colortbl}
\usepackage{tabularx}
\usepackage{multirow}
\usepackage{enumitem}
\usepackage{bbm}
\usepackage{soul}
\usepackage{wrapfig}
\usepackage{cite}
\usepackage{pifont}
\usepackage{booktabs}
\usepackage{natbib}
\usepackage{caption}
\newcolumntype{C}[1]{>{\centering\let\newline\\\arraybackslash\hspace{0pt}}p{#1}}

\def\ie{{\it i.e.}}
\def\eg{{\it e.g.}}

\definecolor{brown}{rgb}{0.65, 0.16, 0.16}
\definecolor{purple}{rgb}{0.54, 0.17, 1.0}
\definecolor{orange}{rgb}{1.0, 0.5, 0.0}
\definecolor{red}{rgb}{1.0, 0.0, 0.0}
\definecolor{green}{rgb}{0, 0.8, 0.5}

\definecolor{citecolor}{HTML}{0071bc}
\usepackage[colorlinks,citecolor=citecolor]{hyperref}

\newcommand{\cmark}{\ding{51}}%
\newcommand{\xmark}{\ding{55}}%

\newcommand{\Fig}[1]{Fig.~\ref{fig:#1}}
\newcommand{\Sec}[1]{Sec.~\ref{sec:#1}}
\newcommand{\Eq}[1]{Eq.~(\ref{eq:#1})}
\newcommand{\Tbl}[1]{Table~\ref{tab:#1}}

\begin{document}

\title{Learning to Detect Semantic Boundaries\\with Image-level Class Labels%
}

\author{Namyup Kim*\thanks{* indicates equal contribution.}\and
        Sehyun Hwang*\and
        Suha Kwak
}

\institute{Namyup Kim \at
              POSTECH, Pohang-si, Republic of Korea \\
              \email{namyup@postech.ac.kr}
           \and
           Sehyun Hwang \at
              POSTECH, Pohang-si, Republic of Korea \\
              \email{sehyun03@postech.ac.kr}
           \and
           Suha Kwak \at
              POSTECH, Pohang-si, Republic of Korea \\
              \email{suha.kwak@postech.ac.kr}
}

\date{Received: 28 August 2020 / Accepted: 19 May 2022}

\maketitle

\begin{abstract}
This paper presents the first attempt to learn semantic boundary detection using image-level class labels as supervision.
Our method starts by estimating coarse areas of object classes through attentions drawn by an image classification network.
Since boundaries will locate somewhere between such areas of different classes, our task is formulated as a multiple instance learning (MIL) problem, where pixels on a line segment connecting areas of two different classes are regarded as a bag of boundary candidates.
Moreover, we design a new neural network architecture that can learn to estimate semantic boundaries reliably even with uncertain supervision given by the MIL strategy.
Our network is used to generate pseudo semantic boundary labels of training images, which are in turn used to train fully supervised models.
The final model trained with our pseudo labels achieves an outstanding performance on the SBD dataset, where it is as competitive as some of previous arts trained with stronger supervision.
\end{abstract}

\section{Introduction}
\label{sec:intro}

Boundary detection has played important roles in many computer vision applications including image and video segmentation~\citep{Arbelaez_TPAMI2011,Galasso_ICCV13}, semantic segmentation~\citep{Banica_CVPR2015,Bertasius_ICCV2015,BNF,domaintransform,Yang_2016_CVPR}, object proposal~\citep{Bertasius_ICCV2015,Yang_2016_CVPR,edge_box}, object detection~\citep{sketch_tokens}, 3D object shape recovery~\citep{Karsch_CVPR2013}, and multi-view stereo~\citep{Shan_CVPR2014}.
Recently, boundary detection performance has been improved substantially thanks to the advance of Convolutional Neural Networks (CNNs)~\citep{HED,DeepContour,DeepEdge,liu2017richer}.
Encouraged by this, the task has been expanded from the conventional binary labeling problem to category-aware detection of boundaries, called Semantic Boundary Detection (SBD)~\citep{Bertasius_ICCV2015,CASENet,SEAL}. 

\begin{figure*} [!t]
\centering
\includegraphics[width = 1 \textwidth]{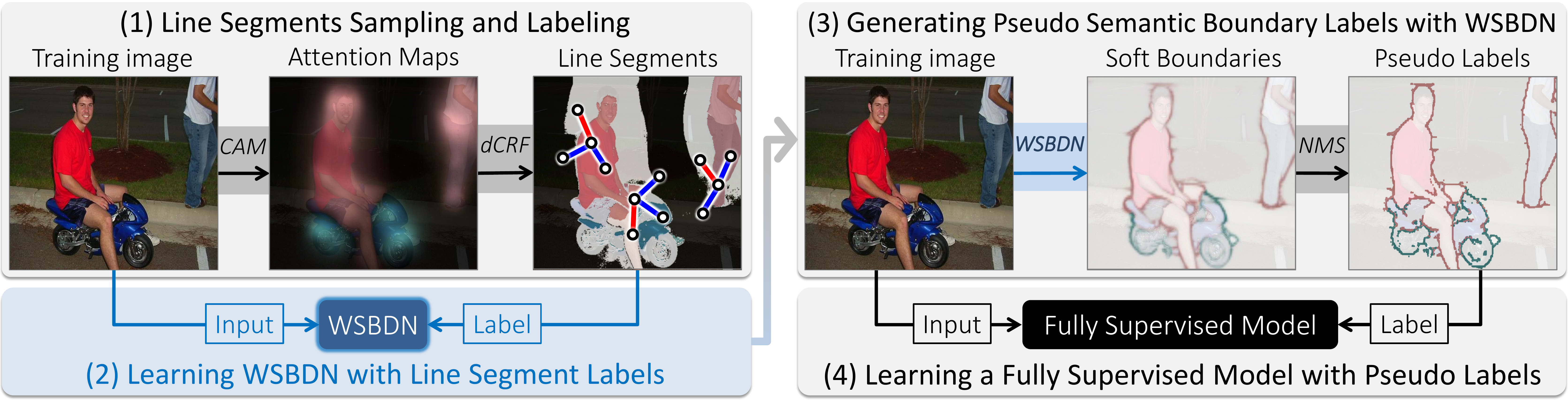}
\caption{
Overview of our approach. 
(1) We first extract line segments and assign semantic boundary labels to them (\Sec{lineseg}).
(2) WSBDN is then trained with the extracted line segments and their labels in a MIL strategy that considers a line segment as a bag of boundary candidates (\Sec{wsbdn}).
(3) The trained WSBDN is applied to training images for generating pseudo semantic boundary labels in a pixel-level (\Sec{fullsup}).
(4) Finally, fully supervised SBD models are trained with the pseudo labels generated by WSBDN (\Sec{fullsup}).
} 
\label{fig:overview}
\end{figure*}

A precondition of the success of CNNs in SBD is a large amount of training images, whose labels are annotated manually in general.
However, since the pixel-level semantic boundary labels are prohibitively expensive,
existing datasets~\citep{cityscapes,SBD} often suffer from the lack of annotated images in terms of quantity and class diversity. 
For this reason, it is not straightforward to learn CNNs that can detect semantic boundaries of various object classes in uncontrolled environments. 

One solution to alleviate this issue is weakly supervised learning, which exploits annotations weaker yet less expensive than semantic boundary labels such as object bounding boxes and image-level class labels.
The low annotation costs of the weak labels allow us to utilize more training images of diverse object classes in the real world.
Especially, image-level class labels are readily available in existing large-scale image classification datasets like ImageNet~\citep{Russakovsky2015}, or can be obtained semi-automatically by web image search where queries are used as class labels of retrieved images.
Thus, the burden of manual annotation can be reduced significantly by adopting image-level class labels as weak supervision.

Weakly supervised learning using image-level class labels has demonstrated successful results in several tasks including semantic segmentation~\citep{affinitynet,sec,Huang_2018_CVPR,oh17cvpr,IRNet,wang2018weakly}, a dual problem of SBD.
To the best of our knowledge, however, there is no existing method for SBD in this line of research, probably because of the inherent difficulty of boundary estimation.
Since boundaries usually have sophisticated structures and occupy only a tiny portion of image pixels, it is hard to infer them given only image-level class labels that indicate nothing but existence of object classes.
What makes the task even harder is the absence of direct cues that compensate for information missing from the weak labels.
In weakly supervised semantic segmentation, for example, most existing methods depend heavily on Class Attention Maps (CAMs)~\citep{Cam,Selvaraju_2017_ICCV,Wei_2017_CVPR} that identify coarse areas of object classes through attention drawn by an image classification CNN.
For semantic boundaries, however, there is no such direct clue that can be obtained without additional supervision.

We propose the first method for learning SBD using image-level class labels as supervision. 
Our method overcomes the aforementioned challenges and is as competitive as previous methods based on stronger supervision.
Our main motivation is that, through CAMs, it is possible to obtain semantic boundary labels reliably in a group-level.
Obviously, CAMs are not accurate enough to estimate semantic boundaries in a pixel-level since their resolution is quite limited and they often capture only local parts of objects.
However, we can at least assume that a group of pixels between different class areas identified in CAMs will include a boundary between the classes.

Our task is thus formulated as a Multiple Instance Learning (MIL) problem~\citep{MIL_Dietterich}, where semantic boundary labels are given in a group-level.
Specifically, we consider the set of pixels on a short line segment as a group, 
and determine semantic boundary labels of the line segment by examining the classes of its two end points in CAMs: If the classes of the end points are different, we consider the line segment contains a boundary pixel as it is crossing a boundary between the two different classes. 
Through MIL with these labeled line segments, our model chooses pixels likely to be semantic boundaries among candidates in the line segments and learns to recognize them simultaneously during training. 
Furthermore, we propose a new CNN architecture, Weakly supervised Semantic Boundary Detection Network (WSBDN), that is designed to capture fine details of boundary shapes as well as to estimate class labels of boundaries accurately in our weakly supervised setting where supervision is uncertain.

Our WSBDN trained through the MIL strategy is then applied to training images for generating their pseudo semantic boundary labels, which are in turn used to learn a fully supervised model for SBD. 
The overall framework of our approach is illustrated in Fig.~\ref{fig:overview}.
The contribution of this paper is three-fold:
\begin{itemize}[leftmargin=5mm] 
\item To the best of our knowledge, our attempt is the first for learning SBD using only image-level class labels as supervision. Furthermore, our method does not rely on external data or off-the-shelf object proposals. 
\item We propose a new CNN architecture, WSBDN, which is well matched with the proposed MIL framework and outperforms existing network architectures for SBD in the weakly supervised setting.
\item Our final model achieves impressive performance on the SBD benchmark~\citep{SBD}, where it is as competitive as previous methods that demand stronger supervision.
\end{itemize}

The remainder of the paper is organized as follows.
\Sec{lineseg} explains our line segment sampling and labeling strategy, and \Sec{wsbdn} describes the architecture of WSBDN and a MIL framework for its training.
Then \Sec{fullsup} describes the process of learning fully supervised SBD models with pseudo semantic boundary labels generated by our model.
In \Sec{experiment}, we analyze the quality of our pseudo labels in depth, and compare our final models with previous arts in SBD and weakly supervised semantic segmentation.

\section{Related Work}
\label{sec:relatedwork}

This section first briefly reviews previous work on boundary detection and SBD, then introduces weakly supervised learning methods closely related to ours.

\subsection{Boundary Detection and SBD}
The goal of boundary detection is to estimate contours of individual objects or surfaces.
As image gradient filters cannot distinguish between such contours and textures, the problem has been addressed by data-driven methods that learn to predict a boundary given a local image patch with ground-truth boundary labels~\citep{Martin_TPAMI04,Dollar_TPAMI2015}.
Motivated by the great success of deep learning, CNNs have been employed for boundary detection.
In early approaches, CNNs take local image patches and estimate the boundary confidence scores of their center locations~\citep{DeepEdge} or their internal boundary configurations directly~\citep{DeepContour}.
Recently, \cite{HED} propose an image-to-image network, called HED, in which all convolution stages of the network are trained to detect boundaries.
Also, \cite{liu2017richer} refine HED by utilizing outputs of all convolution layers and improve the performance by multi-scale augmentation of input image.

SBD is an extension of boundary detection, whose goal is simultaneous detection and classification of boundaries. 
A pioneering work by \cite{SBD} addresses the task by integrating a contour detector with an object detector.
Recently, CNNs have enhanced the performance of SBD substantially.
\cite{Bertasius_ICCV2015} detect class-agnostic boundaries through a boundary detection network and assign class labels to them through a semantic segmentation network.
On the other hand, \cite{CASENet} develop an image-to-image network, called CASENet,
that utilizes lower-level features to enhance high-level boundary classification through shared feature concatenations.
Also, \cite{SEAL} found that ground-truth labels are not accurate enough, and further improve the accuracy of CASENet by optimizing the labels jointly with the network.

All the aforementioned methods are fully supervised and demand semantic boundary labels given manually.
Since such labels are extremely expensive, however, it is not straightforward to learn and deploy the models to handle objects that are not defined in the existing datasets~\citep{cityscapes,SBD}.

\subsection{Weakly Supervised Learning for Vision Tasks}

As a solution to the lack of annotated data in training, weakly supervised learning has become popular in vision tasks for structured prediction such as object detection~\citep{OICR,Bilen16,Kantorov2016,Li2016_weaksup}, semantic segmentation~\citep{IRNet,affinitynet,sec,Huang_2018_CVPR,oh17cvpr,Vernaza2017,Hong2017_webly}, and instance segmentation~\citep{IRNet,PRM,SDI,Cutnpaste} as well as SBD~\citep{wssbdimage,Khoreva_2016_CVPR}.
Since weak labels miss essential details for learning to predict such structured outputs,
the key of success in this line of research is the way of compensating for the missing information during training.
In this context, object box proposals~\citep{SelectiveSearch,random_prim} are used to guess potential locations of objects effectively for object detection and CAMs~\citep{Cam,Selvaraju_2017_ICCV,Wei_2017_CVPR} are employed to estimate coarse areas of object classes for semantic segmentation.
However, there is no such a direct cue for semantic boundaries. 
This makes SBD more difficult than other tasks in weakly supervised learning settings, especially when only image-level class labels are available as supervision.

Previous methods in weakly supervised SBD thus rely on object bounding box labels as supervision.
\cite{Khoreva_2016_CVPR} generate pseudo semantic boundary labels by integrating object boxes with segmentation proposals, and train a fully supervised SBD model with the pseudo labels. 
\cite{wssbdimage} on the other hand tackle the problem by identifying pixels that largely contribute to correct classification of object box images.

Some of recent models for weakly supervised semantic segmentation have a boundary detection capability to improve the quality of their segmentation results by boundary-aware label propagation~\citep{Vernaza2017,IRNet,affinitynet}. 
This capability is learned implicitly~\citep{affinitynet} or explicitly~\citep{IRNet,Vernaza2017} with weak labels such as scribbles~\citep{Vernaza2017} or image-level class labels~\citep{affinitynet,IRNet}.
For example, \cite{affinitynet} train a CNN so that features on its last feature map are similar to each other if their class labels given by CAMs are equal, and detect boundaries by applying a gradient filter to the feature map.
On the other hand, \cite{Vernaza2017} define and optimize a boundary detection network jointly with a segmentation network 
so that the result of the boundary-aware propagation of scribble labels and the output of the segmentation network become identical.
Similarly, the model of \cite{IRNet} has a boundary detection module explicitly, and as in our approach, it is trained through a MIL strategy where a line segment connecting two different class areas discovered by CAMs is considered as a positive bag.

However, the boundary detection modules of these models~\citep{Vernaza2017,IRNet,affinitynet} are not appropriate to address SBD due to the following reasons.
First, they detect class-agnostic boundaries while SBD requires to estimate their class labels as well.
Second, their outputs often highlight non-boundary edges~\citep{IRNet,affinitynet} or cannot capture fine details of boundaries~\citep{Vernaza2017}.
Third, their networks are not optimized for boundary detection or SBD.

Our model is inspired by the above methods~\citep{Vernaza2017,IRNet,affinitynet}, but has clear distinctions.
First, our MIL formulation allows to learn class-aware boundaries.
Second, we propose a new CNN architecture carefully designed for weakly supervised SBD in our setting.
It has been empirically shown that both of our MIL formulation and CNN architecture contribute to the outstanding performance.

\section{Generating Line Segment Labels}
\label{sec:lineseg}

Since pixel-level semantic boundary labels are not available in our setting, we instead utilize line segments and their labels to learn our WSBDN in a MIL framework that considers a line segment as a bag of instances (\ie, pixels).
Formally, a line segment $S_{ij}$ denotes the set of pixels on the line between two image coordinates $\mathbf{x}_i$ and $\mathbf{x}_j$.
Also, a semantic boundary label $c$ is assigned to a line segment $S_{ij}$ if it includes a boundary pixel of class $c$.
Note that a line segment can be assigned no label or multiple labels.

\begin{figure*} [!t]
\centering
\includegraphics[width = 1 \textwidth]{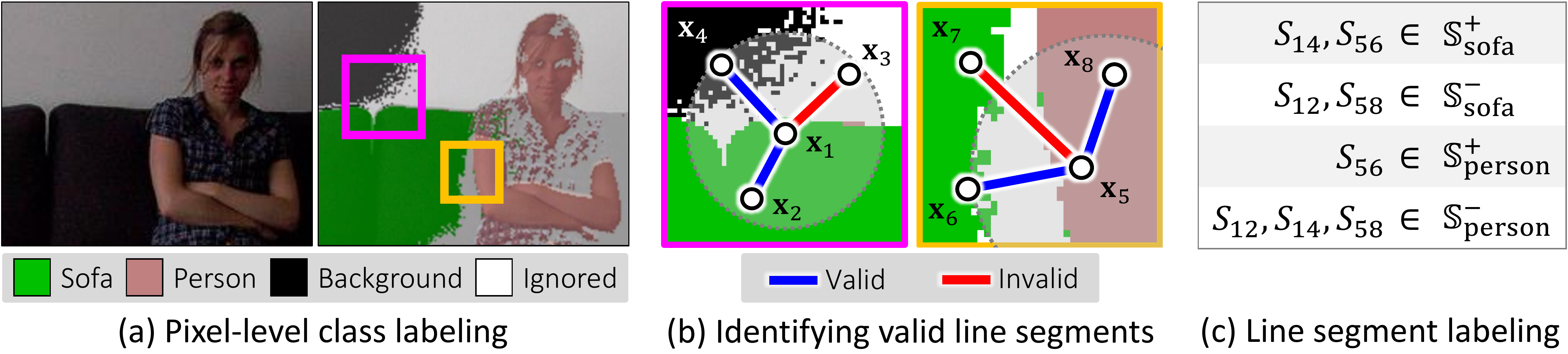}
\caption{
The process of generating line segment labels. 
(a) Pixel-level class labels are first estimated by CAMs and dense CRF~\citep{Fullycrf}. 
(b) Then valid line segments are sampled within a small neighborhood and from only confident regions of objects and background.
(c) Semantic boundary labels of a line segment are determined by the class labels of its two end points and their equivalence. Finally, valid line segments are grouped into multiple sets according to their labels. 
} 
\label{fig:lineseg}
\end{figure*}

One way to estimate semantic boundary labels of $S_{ij}$ is to check the class equivalence between its two end points.
If $\mathbf{x}_i$ and $\mathbf{x}_j$ have different class labels and one of them is $c$, then $c$ is a semantic boundary label of $S_{ij}$ since it is crossing a boundary made by class $c$.
This approach in principle requires pixel-level class labels (\ie, semantic segmentation), which are also not available in our weakly supervised setting.
However, we found that semantic boundary labels of line segments can be estimated reliably by carefully exploiting incomplete semantic segmentation results obtained by CAMs.
The overall process of generating line segment labels is illustrated in Fig.~\ref{fig:lineseg}, and its details are described in the remaining part of this section.

\subsection{Pixel-level Class Labeling by CAMs}
\label{sec:pixel_label}

The line segment labeling process begins by estimating pixel-level class labels through CAMs. 
For computing CAMs, we choose the technique proposed in~\cite{Cam} for its simplicity.
It employs an image classification CNN with a global average pooling followed by the last classification layer. 
Given an image, the CAM of class $c$, denoted by $M_{c}$, is computed by
\begin{equation}
    M_{c} (\mathbf{x}) = \frac{\mathbf{w}_c^\top F(\mathbf{x})}{\max_{\mathbf{x}'} \mathbf{w}_c^\top 
    F(\mathbf{x}')},
\end{equation}
where $\mathbf{x}$ is a coordinate on the last convolution feature map $F$
and $\mathbf{w}_c$ is the classification weight vector for class $c$.
We employ ResNet50~\citep{resnet} to obtain $F$.

$M_{c}$ roughly identifies image regions relevant to class $c$.
However, since $F$ has a limited resolution and $\mathbf{w}_c$ is learned discriminatively, it is blurry and often discovers only local distinctive parts of objects.
To alleviate these issues, we first identify confident regions of object classes and background by thresholding the attention scores conservatively, then refine them through the dense CRF~\citep{Fullycrf}.
Specifically, image regions with top 70\% of attention scores are considered as confident for object classes and those with bottom 5\% are regarded as confident background; regions that turn out to be not confident are ignored.
The output of the refinement is a binary label map $\tilde{M}_c$ for each class $c$, where $\tilde{M}_c(\mathbf{x})=1$ if the image coordinate $\mathbf{x}$ is included in a confident region of class $c$ and $\tilde{M}_c(\mathbf{x})=0$ otherwise.
Qualitative examples of $\tilde{M}_c$ can be found in Fig.~\ref{fig:lineseg}(a).

\subsection{Line Segment Sampling and Labeling}
\label{sec:lineseg_label}

There exist an extremely large number of line segments in an image as they can be sampled by choosing any arbitrary pairs of image coordinates.
Since it is not efficient and effective to take all of them into account during training, we sample and utilize only a small subset of them, called valid line segments.
To be a valid line segment, a pair of sampled coordinates must satisfy the following two conditions: 
(1) The distance between them is smaller than a hyper-parameter $\gamma$, and (2) both of them are sampled from confident regions discovered in \Sec{pixel_label}.
The first condition limits the maximum length of line segments, 
which reduces the computational complexity of training and helps to estimate semantic boundary labels of line segments reliably.
Also, by the second condition, we ignore line segments whose labels cannot be estimated confidently.

Semantic boundary labels of a valid line segment are determined by investigating the pixel-level class label maps obtained in \Sec{pixel_label}.
Specifically, a valid line segment $S_{ij}$ is assigned a semantic boundary label $c$ if the class labels of $\mathbf{x}_i$ and $\mathbf{x}_j$ are different and one of them is $c$ on the label map.
This condition is simply represented by $\tilde{M}_c(\mathbf{x}_i) \neq \tilde{M}_c(\mathbf{x}_j)$.
Thus, the positive set of valid line segments with a semantic boundary label $c$ is denoted and defined by
\begin{align}
\mathbb{S}_c^+ = \big\{S_{ij} \mid \tilde{M}_c(\mathbf{x}_i) \neq \tilde{M}_c(\mathbf{x}_j), S_{ij} \in \mathbb{S} \big\},
\end{align}
where $\mathbb{S}$ is the set of all valid line segments.
We also define the negative set of valid line segments that do not cross a boundary of class $c$ by $\mathbb{S}_c^- = \mathbb{S} - \mathbb{S}_c^+$.
According to the above definition, a line segment can be a member of positive sets of at most two different classes. 
This seems restrictive since a line segment may cross multiple boundaries made by more than two classes, but we argue that one can avoid such a case with a sufficiently small $\gamma$.
The overall procedure for generating line segment labels is illustrated in Fig.~\ref{fig:lineseg}.

\begin{figure*} [!t]
\centering
\includegraphics[width = 1 \textwidth]{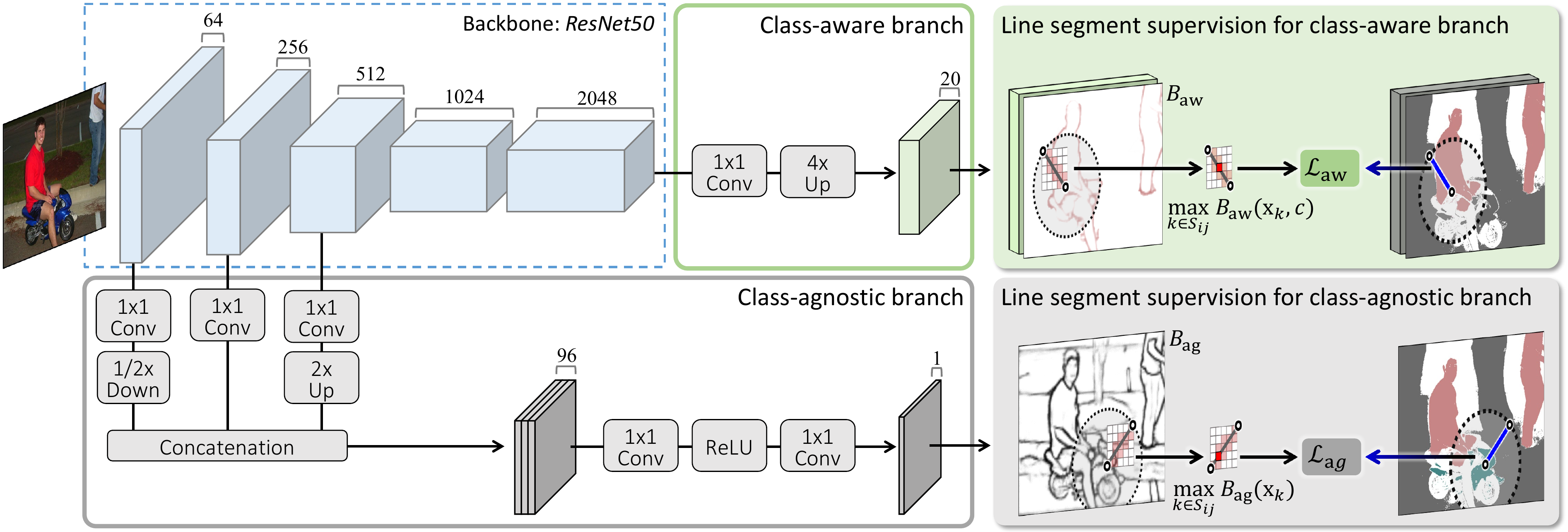}
\caption{
Illustration of the WSBDN architecture and its training strategy.
The final output of WSBDN is obtained by combining its class-agnostic and class-aware boundary maps (\ie, $\mathbf{B}_{\textrm{ag}}$ and $\mathbf{B}_{\textrm{aw}}$) through pixel-wise multiplication.
} 
\label{fig:wsbdn_arch}
\end{figure*}

\section{Weakly Supervised Semantic Boundary Detection Network}
\label{sec:wsbdn}

This section describes details of WSBDN architecture and presents a MIL strategy to train the network with the line segment labels obtained in~\Sec{lineseg}.

\subsection{Architecture}

To detect high-quality semantic boundaries, our model should (1) estimate fine details of boundary shapes accurately and (2) predict class labels of boundaries correctly.
However, it is challenging to achieve both two targets at once in our weakly supervised setting due to the ambiguity of the line segment labels.
To alleviate this issue, we decouple the two targets in our WSBDN architecture, which accordingly has two distinct output branches specialized to each target.
The first branch aims to recognize elaborate shapes of boundaries accurately in a class-agnostic manner.
Meanwhile, the second branch focuses more on predicting class labels of boundaries correctly although their estimated shapes are not crisp. 
The two branches are learned with different losses (\Sec{mil_wsbdn}), and their outputs are combined into the final result by pixel-wise multiplication.

The overall architecture of WSBDN is illustrated in \Fig{wsbdn_arch}.
We employ ResNet50~\citep{resnet} as the backbone network of the two output branches, but slightly modify the original network by decreasing the stride of convolution layers in the last stage\footnote{A stage denotes a group of convolution layers of the same output size.} from 2 to 1 to enlarge the last feature map twice. 
The first branch receives convolution feature maps from the first three stages of the backbone, which are appropriate for predicting elaborate boundary shapes since they capture low-level details of input image in relatively high resolutions. 
The input feature maps are then projected by 1$\times$1 convolution layers and concatenated into a single feature map, which is in turn processed by 1$\times$1 convolution layers with ReLU to predict a single class-agnostic boundary map.
In contrast, the second branch takes the last convolution feature map of the backbone since its large receptive field and rich feature representation enable accurate boundary classification.
The input feature map is fed into the final 1$\times$1 convolution layer that predicts a boundary map for each class.

\begin{figure}[!t]
\begin{center}
\includegraphics[width=0.95 \linewidth]{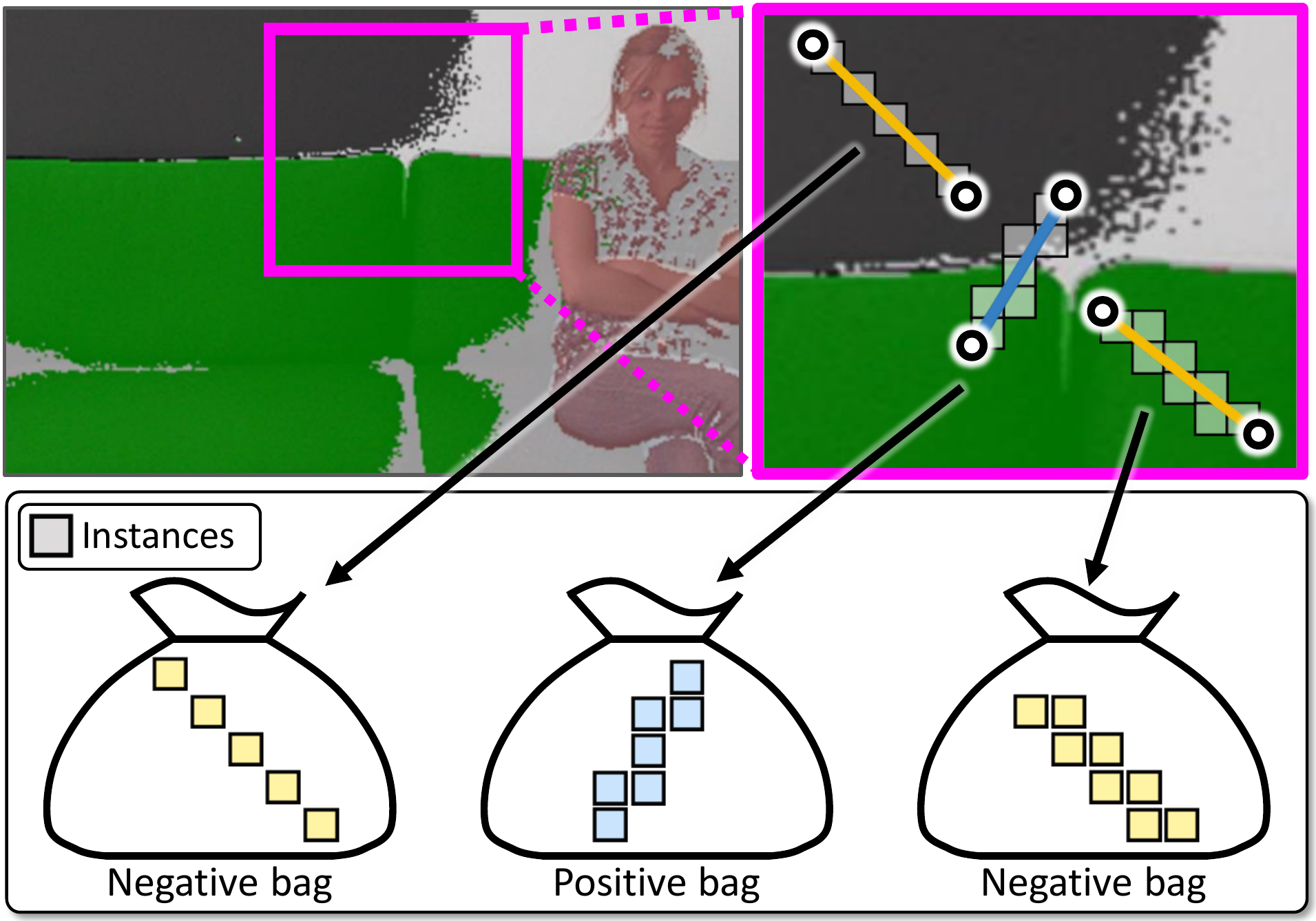}
\end{center}
\vspace{-4mm}
\caption{
Illustration of our MIL framework that considers a line segment as a bag of instances, \ie, pixels on the line segment.
As an aggregation function for MIL, we adopt $\max$ operator.
}
\label{fig:mil}
\end{figure}

\begin{figure*} [!t]
\centering
\includegraphics[width = 0.98 \linewidth]{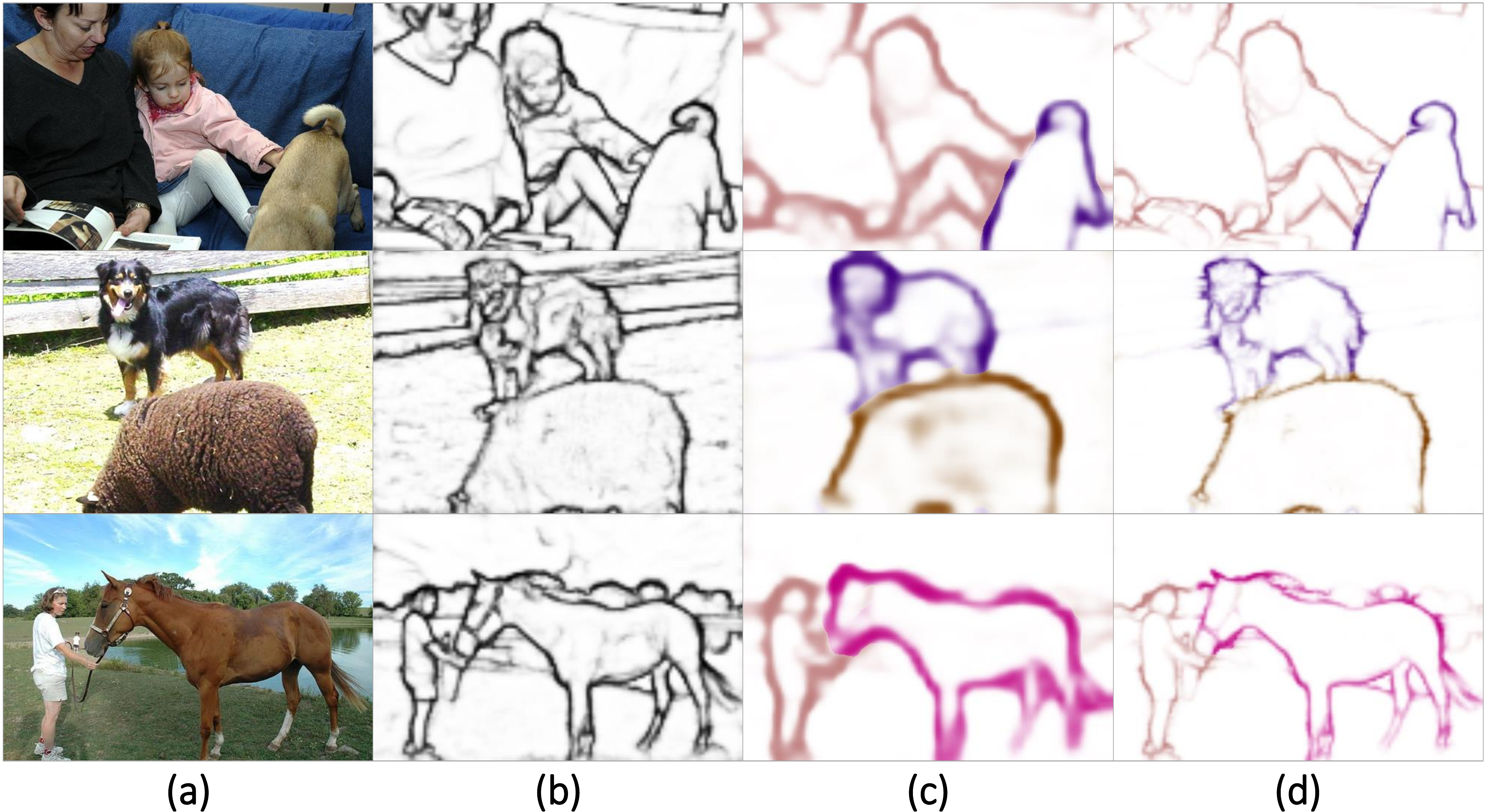}
\caption{
Qualitative results of WSBDN.
(a) Input image. 
(b) Class-agnostic boundary detection. 
(c) Class-aware boundary detection. 
(d) Final output obtained by element-wise multiplication of the two outputs.
} 
\label{fig:wsbdn_outputs}
\end{figure*}

The outputs of the two branches and the final results are compared qualitatively in \Fig{wsbdn_outputs}.
The first branch well captures details of boundary structures, 
but often produces false positives.
The second branch detects object boundaries only and classifies them accurately, but their shapes are usually blurry.
The final output obtained by the pixel level multiplication of the two results is accurate in terms of both shapes and class labels of boundaries.
We argue that the two separate output branches enable WSBDN to achieve the two targets easier thanks to input features and network modules specialized to each of the branches.
Note also that the first branch exploits all positive line segments regardless of their classes (\ie, $\bigcup_{c=1}^C \mathbb{S}^+_c$) for training, which enhances the quality of class-agnostic boundary detection by exploiting a larger number of line segments, and consequently, improves accuracy of the final result as well.

\subsection{Learning WSBDN with Line Segment Labels}
\label{sec:mil_wsbdn}

Image-to-image CNNs like WSBDN are usually supervised by pixel-level labels, which however are not available in our setting. 
We thus propose a MIL strategy that allows to train WSBDN using the line segment labels obtained in \Sec{lineseg};
\Fig{mil} presents a conceptual illustration of instances and bags in our strategy.
We first collect boundary-like pixels by sampling the pixel with the maximum boundary score from each line segment in $\mathbb{S}$;
the $\max$ operation applied here can be interpreted as an aggregation function in the context of MIL.
Also, the class label of the sampled pixel is determined by that of the line segment it belongs to, which means that the pixel is regarded as a boundary pixel if it is a member of a positive line segment, and as a hard negative pixel otherwise. 
WSBDN is then trained to predict the labels of the boundary-like pixels as illustrated in Fig.~\ref{fig:wsbdn_arch}.

Let $\mathbf{B}_{\textrm{ag}} \in \mathbb{R}^{w \times h}$ be the class-agnostic boundary map predicted by the first branch, where $w$$\times$$h$ denotes the output resolution.
The boundary-like pixel $b_{ij}$ of a line segment $S_{ij} \in \mathbb{S}$ is denoted and defined by
\begin{equation}
    b_{ij} = \max \limits_{k \in S_{ij}}\mathbf{B}_{\textrm{ag}}(\mathbf{x}_{k}),
\end{equation}
where $\max(\cdot)$ operation is used as the aggregation function for MIL.
When applying the binary cross entropy loss to each boundary-like pixel, the total loss for $\mathbf{B}_{\textrm{ag}}$ is given by
\begin{equation}
    \mathcal{L}_{\textrm{ag}} = -\sum_{S_{ij} \in \mathbb{S}^{+}} \frac{\log b_{ij}}{| 
        \mathbb{S}^+ |} 
        - \sum_{S_{ij} \in \mathbb{S}^{-}}  \frac{\log(  1 - b_{i j})}{| \mathbb{S}^- |}, 
        \label{eq:loss_ag}
\end{equation}
where $\mathbb{S}^{+} = \bigcup_{c=1}^C \mathbb{S}^+_c$ is the set of positive line segments of all classes, and $\mathbb{S}^{-} = \bigcap_{c=1}^C \mathbb{S}^-_c$ is the set of negative line segments for all classes. Note that $\mathbb{S}^{-} = \mathbb{S} - \mathbb{S}^+$.
In~\Eq{loss_ag}, positive and negative losses are normalized separately since the population of positive and negative line segments is highly imbalanced in general~\citep{CASENet, HED}.

The loss for the second branch is a multi-class extension of $\mathcal{L}_{\textrm{ag}}$.
Let $\mathbf{B}_{\textrm{aw}} \in \mathbb{R}^{w \times h \times C}$ denotes the class-aware boundary map predicted by the second branch, where $C$ indicates the number of predefined object classes.
The boundary-like pixel of a line segment $S_{ij} \in \mathbb{S}$ is defined per class by
\begin{equation}
    b^c_{ij} = \max \limits_{k \in S_{ij}}\mathbf{B}_{\textrm{aw}}(\mathbf{x}_{k}, c),
\end{equation}
and the total loss for $\mathbf{B}_{\textrm{aw}}$ is as follows:
\begin{equation}
        \mathcal{L}_{\textrm{aw}} = - \frac{1}{C} \sum_{c=1}^C \Bigg\{ \sum_{S_{ij} \in \mathbb{S}_c^{+}} \frac{\log b_{ij}^c}{| \mathbb{S}^+_c |} 
        + \sum_{S_{ij} \in \mathbb{S}^-_c}  \frac{\log(  1 - b_{i j}^c)}{| \mathbb{S}^-_c |} \Bigg\}.
        \label{eq:loss_aw}
\end{equation}
The normalization in this loss is done independently for each class to prevent the loss from being biased to a few dominant object classes in an image.
Finally, the total loss for WSBDN is given by
\begin{equation}
        \mathcal{L}  =  \mathcal{L}_{\textrm{aw}} + \lambda\mathcal{L}_{\textrm{ag}},
    \label{eq:total_loss}
\end{equation}
where $\lambda$ is a weight hyper-parameter.

\begin{figure*} [!t]
\centering
\includegraphics[width = 1 \textwidth]{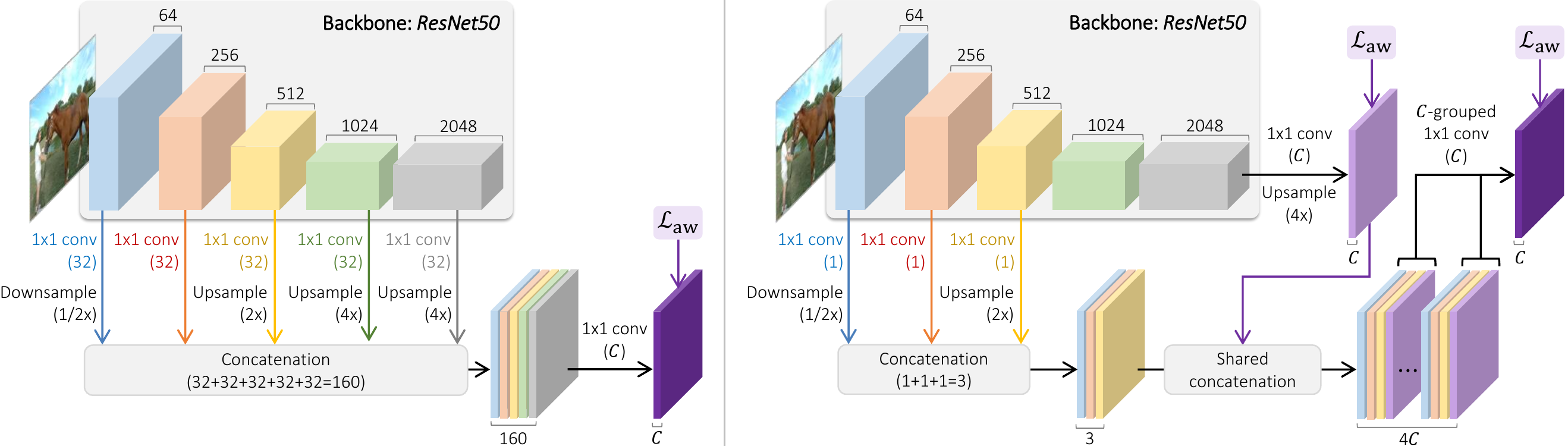}
\caption{
Architectures of Class-aware IRNet (\emph{left}) and WS-CASENet (\emph{right}).  
} 
\label{fig:other_arch}
\end{figure*}

\begin{figure*} [!t]
\centering
\includegraphics[width = 1.0 \textwidth]{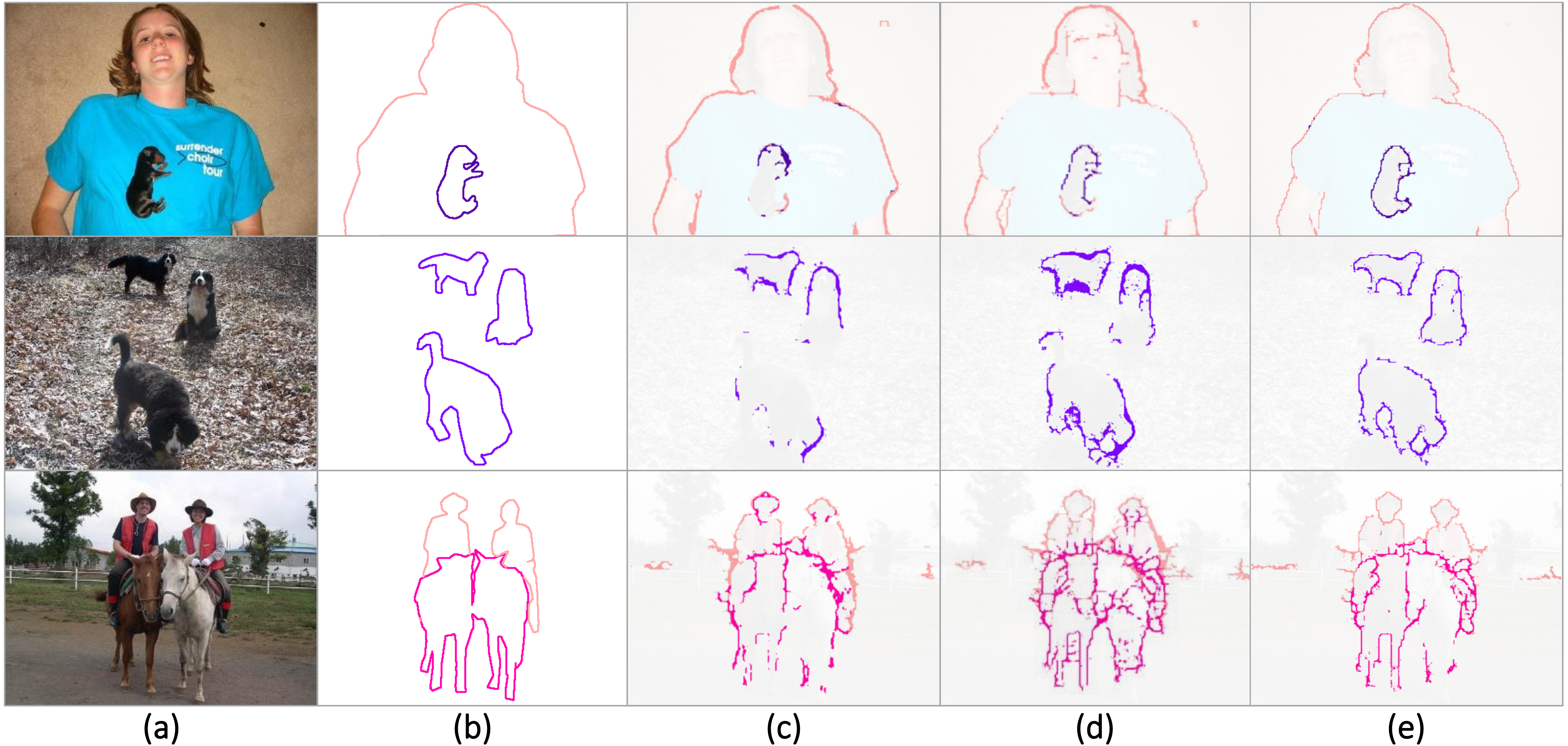}
\caption{
Qualitative comparisons of pseudo semantic boundary labels. 
(a) Input image.
(b) Ground-truth. 
(c) Class-aware IRNet. 
(d) WS-CASENet.
(e) WSBDN.
} 
\label{fig:qual_pseudo}
\end{figure*}

\section{Learning a Fully Supervised SBD Network}
\label{sec:fullsup}

Following the convention in weakly supervised learning for visual recognition~\citep{affinitynet,IRNet,oh17cvpr,OICR,Hong2017_webly},
the trained WSBDN is applied to training images for generating their pseudo semantic boundary labels, which are used to train fully supervised models for SBD.
We choose CASENet~\citep{CASENet} as the fully supervised model, but any other networks proposed for SBD can be employed.
Note that, unlike the original CASENet whose backbone network is pretrained on the MS-COCO dataset~\citep{Mscoco}, we employ ImageNet pretrained models as our backbone so that the entire framework stays weakly supervised. 

To enhance the quality of the pseudo labels, the three steps listed below are conducted in order. 
First, input image for WSBDN is augmented with multiple scales and horizontal flip, and the outputs are aggregated into a single semantic boundary map.
Second, we filter out boundary scores of the classes irrelevant to the ground-truth image-level class labels. 
Finally, the Non-Maximum Suppression~\citep{canny} integrated with a non-parametric thresholding technique~\citep{otsu1979threshold} is applied to binarize the boundary map.

The main advantage of adopting the above two stage approach, \ie, first generating pseudo labels then training fully supervised models with the pseudo labels, is performance improvement.
Since this approach applies the pseudo-label generator (\ie, WSBDN) to training images whose ground-truth image-level class labels are known, it has a chance to filter out pseudo boundary labels of irrelevant classes.
The final model thus can be trained with pseudo labels whose quality is better than that of the direct predictions of the pseudo label generator.

\section{Experiments}
\label{sec:experiment}

In this section, we first demonstrate the effectiveness of our method on the SBD dataset~\citep{SBD} by evaluating the quality of our pseudo labels and the performance of CASENet trained with them.
In addition, since semantic segmentation is a dual problem of SBD, our model is compared to state-of-the-art weakly supervised semantic segmentation methods relying on the same level of supervision.

\begin{table}[!t] \small
\caption{The quality of pseudo semantic boundary labels evaluated on the SBD benchmark \emph{train} set in mAP (\%) and MF (\%).
\emph{Soft boundary}: Raw outputs predicted by the models.
\emph{Hard boundary}: Pseudo labels generated by the processes in~\Sec{fullsup}.
}
\centering
\begin{tabular}{lcccc}
    \toprule
    \multirow{2}{*}{}& \multicolumn{2}{c}{Soft boundary} & \multicolumn{2}{c}{Hard boundary}\\
    & mAP  & MF & mAP & MF \\
    \midrule
    Class-aware IRNet & 55.7 & 59.7 & 35.7 & 58.0 \\
    WS-CASENet & 50.1 & 59.2 & 33.5 & 56.0 \\
     WSBDN (Ours)  & \textbf{58.0} & \textbf{62.4} & \textbf{38.1} & \textbf{59.8} \\
    \bottomrule
\end{tabular}
\label{tab:pseudo_label_accu}
\end{table}

\begin{figure} [!t]
\includegraphics[width = 1.0 \linewidth]{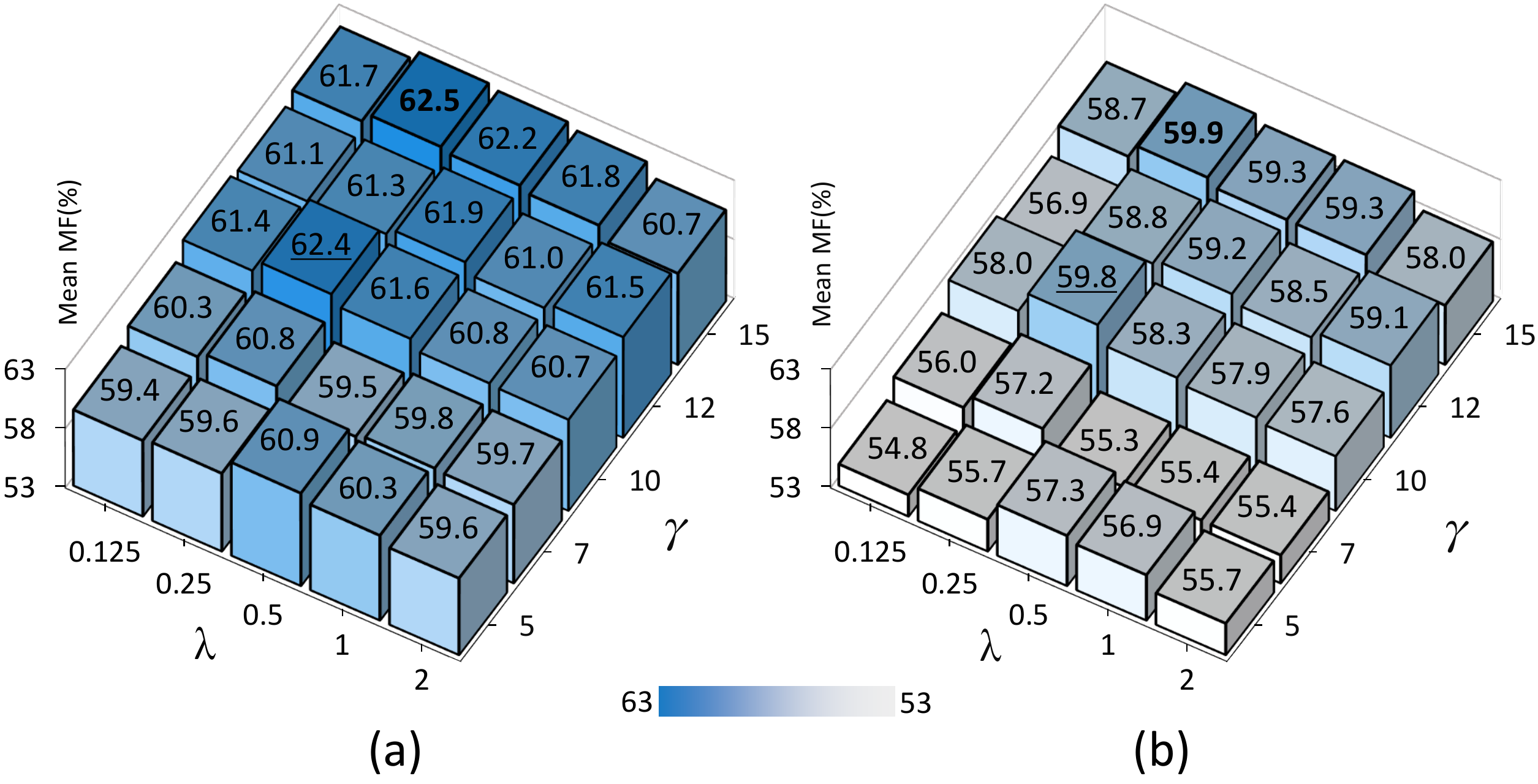}
\caption{The quality of pseudo boundary labels in mean MF (\%) versus the two hyper-parameters $\gamma$ and $\lambda$.
(a) Soft boundary. (b) Hard Boundary.
The underlined score is obtained by our parameter setting while the bold score is the optimal one in this grid search.
}
\label{fig:effect_hyperparam}
\end{figure}

\subsection{Implementation Details}
\noindent \textbf{Dataset and evaluation protocol.}
Experiments have been done in the SBD dataset~\citep{SBD}, which consists of 11,355 images from PASCAL VOC~\citep{Pascalvoc}.
Following the \emph{train-test} split in~\cite{CASENet}, 8,498 images among them are used for training and 2,857 images are kept for testing.
Also, following the official evaluation protocol of the dataset, two metrics are used to measure the quality of predicted semantic boundaries: Maximum F-measure (MF) at optimal dataset scale and Average Precision (AP) for each of classes.

\vspace{1mm}
\noindent \textbf{Hyper-parameter setting.} 
For the dense CRF in \Sec{pixel_label}, we use the default parameters given in the original code.
$\gamma$ in \Sec{lineseg_label} and $\lambda$ in \Eq{total_loss} are set to 10 and 0.25, respectively.
The NMS~in \Sec{fullsup} is conducted with suppression radius 10 and conservative multiplier 1.1.

\begin{figure}[!t]
\begin{center}
\includegraphics[width=1 \linewidth]{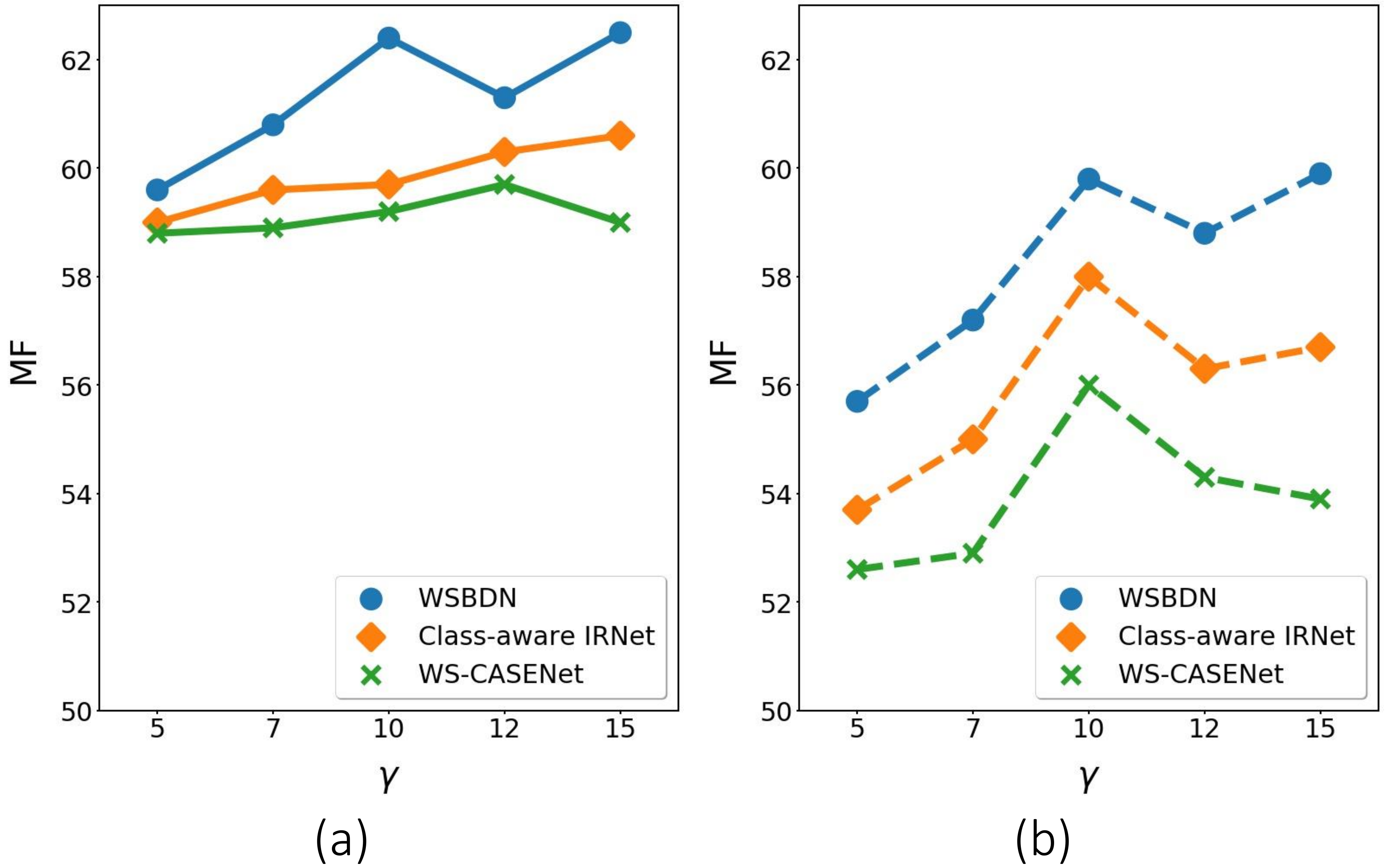}
\end{center}
\caption{The quality of pseudo boundary labels in mean MF (\%) versus the hyper-parameter $\gamma$ on the SBD benchmark \emph{train} set. 
(a) Soft boundary map. 
(b) Hard boundary map. 
} 
\label{fig:gird_plot}
\end{figure}

\vspace{1mm}
\noindent{\textbf{Complexity of identifying valid line segment.}}
The time complexity of identifying valid line segments is $O(C N\gamma^{2})$, where $C$ and $N$ indicate the number of classes and the size of the boundary maps, respectively. 
In practice, this process takes little time as it is fully operated on GPU, and does not affect the inference time of WSBDN as it is done only for training the network.

\vspace{1mm}
\noindent \textbf{Training WSBDN.} 
The backbone of WSBDN is pretrained on the ImageNet~\citep{Imagenet} and its two output branches are randomly initialized. 
WSBDN is then optimized by SGD with polynomial decay, where the initial learning rate, momentum, and weight decay are set to $1e$$-$$2$, 0.9, and $1e$$-$$4$, respectively.

\vspace{1mm}
\noindent \textbf{Training our final model.} 
\iftrue
For CASENet trained with our pseudo labels, we employ various backbone networks, VGG16 and VGG19~\citep{vggnet} with batch normalization~\citep{Batchnorm}, and ResNet50 and ResNet101~\citep{resnet};
unlike that of the original CASENet, these backbone networks are all pretrained on the ImageNet.
The initial learning rate is set to $3e$$-$$2$ and $5e$$-$$2$ for VGG16 and ResNet backbone networks, respectively.
Other details follow the setting presented in~\cite{dff}.
\fi

\subsection{Analysis of Pseudo Semantic Boundary Labels}
\noindent \textbf{Baselines and performance evaluation.}
The quality of pseudo semantic boundary labels generated by our method is evaluated quantitatively and qualitatively on the SBD benchmark \emph{train} set.
Also, to validate the advantage of WSBDN, we compare our pseudo labels with those generated by other network architectures.
We choose IRNet~\citep{IRNet} and CASENet~\citep{CASENet} for the comparisons since IRNet has a internal module for weakly supervised boundary detection, a class-agnostic version of our task, and CASENet is one of the current state-of-the-art architectures for SBD.

\begin{table}[!t] \small
\caption{
The quality of pseudo semantic boundary labels evaluated on the SBD benchmark \emph{train} set in mAP (\%) and MF (\%).
$\mathbf{B}_\textrm{ag}$ and $\mathbf{B}_\textrm{aw}$ indicates the class-agnostic and class-aware boundary maps, respectively, and $\mathbf{B}_\textrm{aw} \odot \mathbf{B}_\textrm{ag}$ denotes the final output obtained by their element-wise multiplication.
\emph{Class-aware}: Multi-class boundary detection. \emph{Class-agnostic}: Binary boundary detection.
}

\centering
\begin{tabular}{lcccc}
    \toprule
    \multirow{2}{*}{}& \multicolumn{2}{c}{Class-aware} & \multicolumn{2}{c}{Class-agnostic}\\
    & mAP  & MF & AP & MF \\
    \midrule
    $\mathbf{B}_\textrm{aw} $ & 46.6 & 60.0 & 53.4 & 65.7 \\
    $\mathbf{B}_\textrm{ag} $ & - & - & 53.6 & 58.1 \\
    $\mathbf{B}_\textrm{aw} \odot \mathbf{B}_\textrm{ag}$ & \textbf{58.0} & \textbf{62.4} & \textbf{66.0} & \textbf{68.2} \\
    \bottomrule
\end{tabular}
\label{tab:branch_pseudo_accu}
\end{table}

Since IRNet predicts class-agnostic boundaries, it cannot be directly employed to generate pseudo semantic boundary labels. 
We thus modify IRNet slightly by increasing the number of output channels of the last prediction layer from 1 to $C$, the number of predefined object classes, and train the model with the class-aware MIL loss in~\Eq{loss_aw}.
We call this modified IRNet \emph{Class-aware IRNet}.
In contrast, CASENet is used without architecture modification since it is designed for SBD, and is trained directly with the class-aware MIL loss in~\Eq{loss_aw}.
We call this weakly supervised version of CASENet \emph{WS-CASENet}.
For fair comparisons, both of Class-aware IRNet and WS-CASENet are based on the same backbone with WSBDN.
The architectures of the two networks are described in~\Fig{other_arch}.

As summarized in~\Tbl{pseudo_label_accu}, WSBDN consistently outperforms Class-aware IRNet and WS-CASENet in both of mAP and MF.
The same tendency can be observed from the qualitative examples in~\Fig{qual_pseudo}, where WSBDN detects the entire object boundaries better than other two networks that often miss non-discriminative parts of object.
Also, WSBDN correctly classifies boundaries of multiple classes while Class-aware IRNet and WS-CASENet are sometimes confused by co-occurring patterns of different classes, \eg, human faces and horses in the third column of~\Fig{qual_pseudo}.

\vspace{1mm}
\noindent{\textbf{Effect of combining $\mathbf{B}_{\textrm{aw}}$ and $\mathbf{B}_\textrm{ag}$.}}
We investigate contributions of the two outputs of WSBDN, $\mathbf{B}_{\textrm{aw}}$ and $\mathbf{B}_\textrm{ag}$, and the effect of combining them by element-wise multiplication.
Due to the absence of class information in $\mathbf{B}_{\textrm{ag}} \in \mathbb{R}^{w \times h}$, its performance is evaluated only for class-agnostic boundary (\ie, binary boundary) detection.
Meanwhile, $\mathbf{B}_{\textrm{aw}}$ and the combined prediction denoted by $\mathbf{B}_\textrm{aw}\odot\mathbf{B}_\textrm{ag}$ are evaluated in both class-wise and class-agnostic manners.
As summarized in~\Tbl{branch_pseudo_accu}, 
$\mathbf{B}_\textrm{aw}\odot\mathbf{B}_\textrm{ag}$ outperforms $\mathbf{B}_\textrm{aw}$ and $\mathbf{B}_\textrm{aw}$ substantially in both class-aware and class-agnostic settings. 
The gaps between $\mathbf{B}_\textrm{aw}\odot\mathbf{B}_\textrm{ag}$ and $\mathbf{B}_\textrm{aw}$ indicate that $\mathbf{B}_\textrm{ag}$ allows to capture fine details of boundaries complementary to $\mathbf{B}_\textrm{aw}$.
Also, $\mathbf{B}_\textrm{aw}\odot\mathbf{B}_\textrm{ag}$ improves performance of $\mathbf{B}_\textrm{ag}$ even in the class-agnostic setting since $\mathbf{B}_\textrm{aw}$ is useful to filter out false positive boundaries frequent in $\mathbf{B}_\textrm{ag}$.
These results clearly justify the use of $\mathbf{B}_\textrm{aw}\odot\mathbf{B}_\textrm{ag}$ as the final output of WSBDN, and demonstrate that both of $\mathbf{B}_\textrm{aw}$ and $\mathbf{B}_\textrm{ag}$ contribute to the quality of the final output.

\begin{figure}[!t]
\begin{center}
\includegraphics[width=0.95\linewidth]{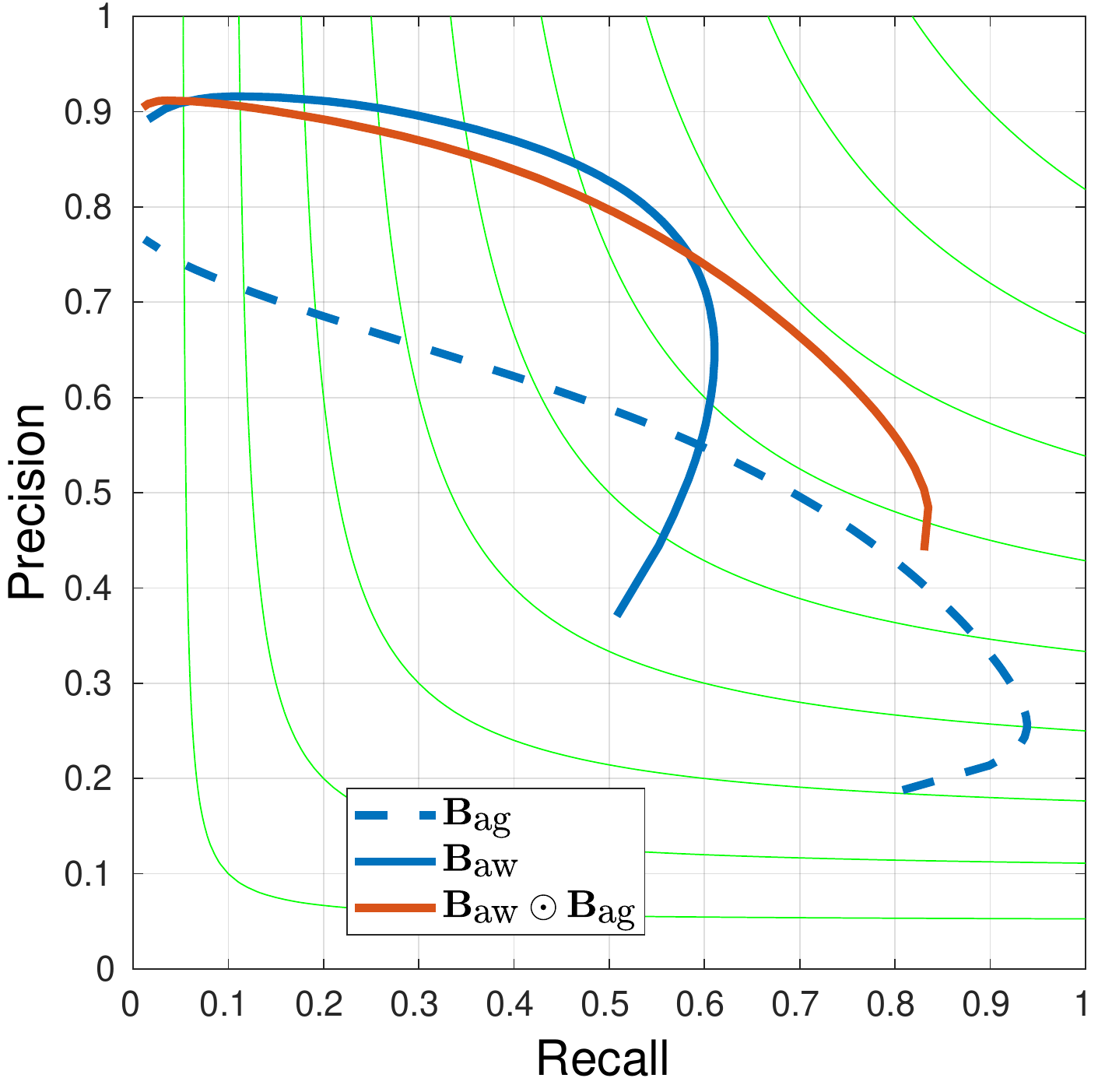}
\end{center}
\vspace{-6mm}
\caption{
PR curves of class-agnostic boundary detection of $\mathbf{B}_{\textrm{aw}}$, $\mathbf{B}_{\textrm{ag}}$, and their combination $\mathbf{B}_\textrm{aw}\odot\mathbf{B}_\textrm{ag}$.
}
\label{fig:bin_pr_curve}
\end{figure}

To further clarify the roles of $\mathbf{B}_{\textrm{aw}}$ and $\mathbf{B}_{\textrm{ag}}$, we draw PR curves of their class-agnostic boundary detection using the official SBD evaluation protocol~\citep{SBD}.
As shown in~\Fig{bin_pr_curve}, both AP and MF are improved by combining $\mathbf{B}_{\textrm{aw}}$ with higher precision and $\mathbf{B}_\textrm{ag}$ with higher recall.

\begin{figure} [!t]
\includegraphics[width = 1.0 \linewidth]{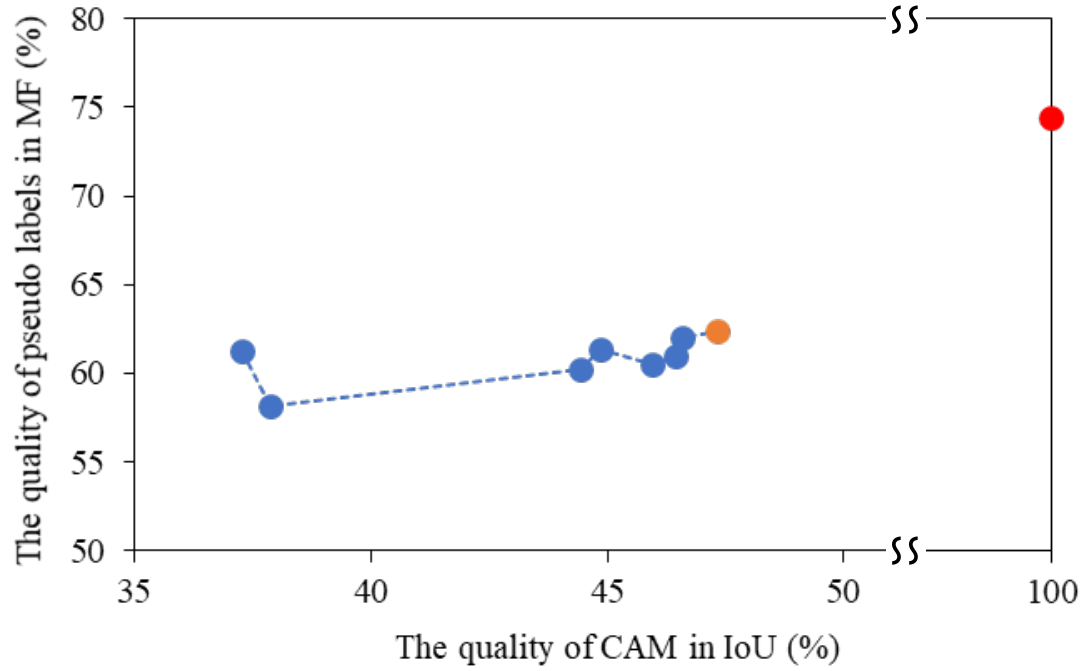}
\caption{
The quality of pseudo boundary labels in mean MF (\%) versus quality of CAM in IoU (\%).
The orange dot indicates the CAM adopted in our method
and the red dot means using ground-truth segmentation labels instead of CAM.
}
\label{fig:effect_camiou}
\end{figure}

\vspace{1mm}
\noindent{\textbf{Effect of the performance of CAM.}}
Since our pseudo labels are generated from CAM, the performance of CAM may affect the performance of WSBDN.
To demonstrate the robustness of WSBDN to errors in CAMs, we analyze the performance of WSBDN while varying the quality of CAMs used to train it.
Specifically, diverse and deteriorated CAMs are generated by under-fitted classification networks in the early stages of training.
As summarized in~\Fig{effect_camiou}, the accuracy of pseudo labels generated by WSBDN is fairly stable even with severely deteriorated CAMs, \eg, when IoU of CAM drops by 10\%, MF score of the pseudo boundary labels drops by 4.5\%.
This robustness comes mainly from our line segment labeling strategy that considers only confident areas of CAMs.
Additionally, we conduct an upper-bound test by investigating performance of WSBDN when using ground-truth segmentation masks instead of CAM to generate line segment labels.
As shown in~\Fig{effect_camiou}, the performance of WSBDN is significantly improved, up to 74.4\%, by using the ground-truth masks.
This results demonstrate the potential of WSBDN; it can be enhanced a lot by improving the quality of CAM.

\vspace{1mm}
\noindent{\textbf{Effect of the hyper-parameters.}}
We investigate the effect of $\gamma$ and $\lambda$ on the quality of pseudo labels.
As shown in~\Fig{effect_hyperparam}, the quality is fairly stable, varying less than 3\% in MF, when $\gamma\geq10$.
Note that $\gamma=10$ and $\lambda=0.25$ used in our paper are not optimal; we set them based on our common sense as there is no perfect way to optimize them in the weakly supervised setting with no boundary label.
In addition, we investigate the impact of the hyper-parameter $\gamma$ on the quality of pseudo semantic boundary labels generated by the three different architectures: WSBDN, Class-aware IRNet, and WS-CASENet. 
The other hyper-parameter $\lambda$ is fixed to 0.25 for ease of analysis.
As shown in~\Fig{gird_plot}, the quality score is fairly stable when $\gamma\geq10$ in the case of WSBDN.
More importantly, WSBDN consistently outperforms Class-aware IRNet and WS-CASENet for all values of $\gamma$ we examined.
This demonstrates the superiority of our WSBDN over the other architectures in pseudo boundary label generation.

\begin{table}[t!]
\centering
\caption{The quality of pseudo semantic boundary labels evaluated on the SBD benchmark \emph{train} set in mAP (\%) and MF (\%). MSF: Multi-scales and horizontal flips. NMS: Non-maximum suppression.
}
\begin{tabular}{cc|cccc}
\toprule
    &     & \multicolumn{2}{c}{Soft boundary}                  & \multicolumn{2}{c}{Hard boundary} \\
MSF & NMS & mAP               & MF                    & mAP        & MF          \\ \midrule
 \multirow{2}{*}{\xmark}   & \xmark   & \multirow{2}{*}{57.7} & \multirow{2}{*}{61.9} & 35.5 & 56.5        \\
  & \cmark   &                       &                       & 34.5 & 55.5        \\\midrule
 \multirow{2}{*}{\cmark}   & \xmark   & \multirow{2}{*}{58.0} & \multirow{2}{*}{62.4} & 36.1 & 57.7        \\
  & \cmark   &                       &                       & 38.1 & 59.8        \\
\bottomrule
\end{tabular}
\label{tab:msf_nms}
\end{table}

\vspace{1mm}
\noindent{\textbf{Effect of the post-processing.}}
We investigate the effect of the post-processing, \ie, the inference-time augmentation by Multiple Scales and horizontal Flips (MSF) and Non-Maximum Suppression (NMS).
As shown in~\Tbl{msf_nms}, the improvement by MSF is less than 1\% in all settings, and NMS improves the quality of our pseudo labels by about 2\% only when combined with MSF.
These results demonstrate that WSBDN does not heavily depend on the post-processing techniques, although they help improve the quality of our pseudo labels slightly.

\begin{table*}[!t]
\vspace{-2mm}
\caption{Performance on the SBD benchmark \emph{test} set in MF (\%). 
The backbone networks of the methods (B.) are denoted by
$\mathcal{A}$--AlexNet, 
$\mathcal{V}_{16}$--VGG16, 
$\mathcal{V}_{19}$--VGG19,
$\mathcal{R}_{50}$--ResNet50, and
$\mathcal{R}_{101}$--ResNet101.
Also, the type of supervision (S.) indicates $\mathcal{F}$--pixel-level semantic boundary, $\mathcal{B}$--object bounding box, and $\mathcal{I}$--image-level class label.
DFF~\citep{dff} reproduced by the official implementation is indicated by $\dagger$.
}
\vspace{-2mm}
\centering
\scalebox{0.88}{
\begin{tabular}
{
@{}C{3.2cm}@{}@{}C{0.9cm}@{}C{0.5cm}@{}|@{}C{0.7cm}@{}@{}C{0.7cm}@{}@{}C{0.7cm}@{}@{}C{0.7cm}@{}@{}C{0.7cm}@{}@{}C{0.7cm}@{}@{}C{0.7cm}@{}@{}C{0.7cm}@{}@{}C{0.7cm}@{}@{}C{0.7cm}@{}@{}C{0.7cm}@{}@{}C{0.7cm}@{}@{}C{0.7cm}@{}@{}C{0.7cm}@{}@{}C{0.7cm}@{}@{}C{0.7cm}@{}@{}C{0.7cm}@{}@{}C{0.7cm}@{}@{}C{0.7cm}@{}@{}C{0.7cm}@{}|@{}C{0.8cm}@{}
}
\toprule
Method\raggedright& B. &S.&aer&bik&bir&boa&bot&bus&car&cat&cha&cow&tab&dog&hor&mbk&per&pla&she&sof&trn&tv&mean\\
\midrule
\cite{SBD}\raggedright & - & $\mathcal{F}$ & 41.5 & 46.7 & 15.6 & 17.1 & 36.5 & 42.6 & 40.3 & 22.7 & 18.9 & 26.9 & 12.5 & 18.2 & 35.4 & 29.4 & 48.2 & 13.9 & 26.9 & 11.1 & 21.9 & 31.4 & 27.9\\ 
\cite{Bertasius_ICCV2015}\raggedright & $\mathcal{V}_{19}$ & $\mathcal{F}$ & 71.6 & 59.6 & 68.0 & 54.1 & 57.2 & 68.0 & 58.8 & 69.3 & 43.3 & 65.8 & 33.3 & 67.9 & 67.5 & 62.2 & 69.0 & 43.8 & 68.5 & 33.9 & 57.7 & 54.8 & 58.7\\
\cite{Bertasius_ICCV2015}\raggedright & $\mathcal{V}_{19}$ & $\mathcal{F}$  & 73.9 & 61.4 & 74.6 & 57.2 & 58.8 & 70.4 & 61.6 & 71.9 & 46.5 & 72.3 & 36.2 & 71.1 & 73.0 & 68.1 & 70.3 & 44.4 & 73.2 & 42.6 & 62.4 & 60.1 & 62.5\\ 
\cite{CASENet}\raggedright & $\mathcal{V}_{19}$ & $\mathcal{F}$  & 72.5 & 61.5 & 63.8 & 54.5 & 52.3 & 65.4 & 62.6 & 67.2 & 42.6 & 51.8 & 31.4 & 62.0 & 61.9 & 62.8 & 75.4 & 41.7 & 59.8 & 35.8 & 59.7 & 50.7 & 56.8\\
\cite{CASENet}\raggedright & $\mathcal{R}_{101}$ & $\mathcal{F}$  & 83.3 & 76.0 & 80.7 & 63.4 & 69.2 & 81.3 & 74.9 & 83.2 & 54.3 & 74.8 & 46.4 & 80.3 & 80.2 & 76.6 & 80.8 & 53.3 & 77.2 & 50.1 & 75.9 & 66.8 & 71.4\\
\cite{SEAL}\raggedright & $\mathcal{R}_{101}$ & $\mathcal{F}$ & 84.5 & 76.5 & 83.7 & 64.9 & 71.7 & 83.8 & 78.1 & 85.0 & 58.8 & 76.6 & 50.9 & 82.4 & 82.2 & 77.1 & 83.0 & 55.1 & 78.4 & 54.4 & 79.3 & 69.6 & 73.8\\
\cite{dff}\raggedright & $\mathcal{R}_{101}$ & $\mathcal{F}$   & 86.5 & 79.5 & 85.5 & 69.0 & 73.9 & 86.1 & 80.3 & 85.3 & 58.5 & 80.1 & 47.3 & 82.5 & 85.7 & 78.5 & 83.4 & 57.9 & 81.2 & 53.0 & 81.4 & 71.6 & 75.4\\
\cite{dff}$^\dagger$\raggedright & $\mathcal{R}_{101}$ & $\mathcal{F}$   & 84.7 & 76.8 & 83.5 & 67.2 & 72.7 & 84.4 & 77.7 & 86.7 & 58.3 & 77.6 & 51.7 & 83.4 & 84.0 & 67.5 & 86.1 & 57.6 & 75.5 & 53.2 & 80.3 & 72.2 & 74.1\\
\cite{Acuna_2019_CVPR}\raggedright & $\mathcal{R}_{101}$ & $\mathcal{F}$   & 85.8 & 80.0 & 85.6 & 68.4 & 71.6 & 85.7 & 78.1 & 87.5 & 59.1 & 78.5 & 53.7 & 84.8 & 83.4 & 79.5 & 85.3 & 60.2 & 79.6 & 53.7 & 80.3 & 71.4 & 75.6\\
\midrule
\cite{wssbdimage}\raggedright & $\mathcal{A}$ & $\mathcal{B}$ & - & - & - & - & - & - & - & - & - & - & - & - & - & - & - & - & - & - & - & - & 38.1\\
\cite{Khoreva_2016_CVPR}\raggedright & $\mathcal{V}_{16}$ & $\mathcal{B}$ & 65.9 & 54.1 & 63.9 & 47.9 & 47.0 & 60.4 & 50.9 & 56.5 & 40.4 & 56.0 & 30.0 & 57.5 & 58.0 & 57.4 & 59.5 & 39.0 & 64.2 & 35.4 & 51.0 & 42.4 & 51.9\\
\midrule
Ours+CASENet \raggedright & $\mathcal{V}_{16}$ & $\mathcal{I}$ & 67.9 & 59.8 & 59.5 & 49.3 & 44.3 & 57.7 & 52.1 & 63.5 & 30.7 & 46.8 & 22.6 & 56.5 & 58.4 & 55.4 & 65.1 & 40.6 & 58.8 & 29.6 & 43.4 & 45.1 & 50.2\\
Ours+CASENet \raggedright & {$\mathcal{V}_{19}$} & {$\mathcal{I}$} & {68.5} & {62.7} & {61.9} & {47.3} & {45.8} & {61.2} & {54.4} & {65.0} & {31.9} & {49.1} & {24.9} & {58.5} & {61.8} & {58.8} & {65.7} & {41.3} & {61.2} & {31.2} & {44.8} & {45.9} & {52.1}\\
Ours+CASENet\raggedright & $\mathcal{R}_{50}$ & $\mathcal{I}$ & 73.4 & 69.3 & 69.9 & 52.2 & 55.7 & 69.2 & 57.8 & 74.4 & 37.6 & 66.6 & 31.7 & 71.7 & 73.1 & 68.8 & 65.9 & 47.5 & 71.9 & 38.3 & 51.0 & 50.7 & 59.8\\
Ours+CASENet\raggedright & $\mathcal{R}_{101}$ & $\mathcal{I}$ & 74.1 & 67.7 & 70.6 & 52.4 & 56.1 & 67.3 & 57.6 & 73.3 & 38.7 & 69.1 & 31.5 & 70.6 & 73.2 & 66.3 & 67.6 & 46.8 & 71.8 & 38.7 & 50.0 & 50.2 & 59.7 \\
Ours+DFF$^\dagger$\raggedright & $\mathcal{R}_{101}$ & $\mathcal{I}$ & 74.6 & 68.2 & 71.5 & 52.5 & 58.9 & 67.1 & 59.2 & 74.6 & 39.9 & 68.8 & 33.2 & 72.6 & 74.6 & 67.4 & 68.3 & 48.3 & 73.3 & 39.1 & 49.2 & 53.1 & \textbf{60.7} \\
\bottomrule
\end{tabular}}
\label{tab:sbd_test}
\end{table*}

\begin{figure*} [!t]
\centering
\includegraphics[width = 1.0 \textwidth]{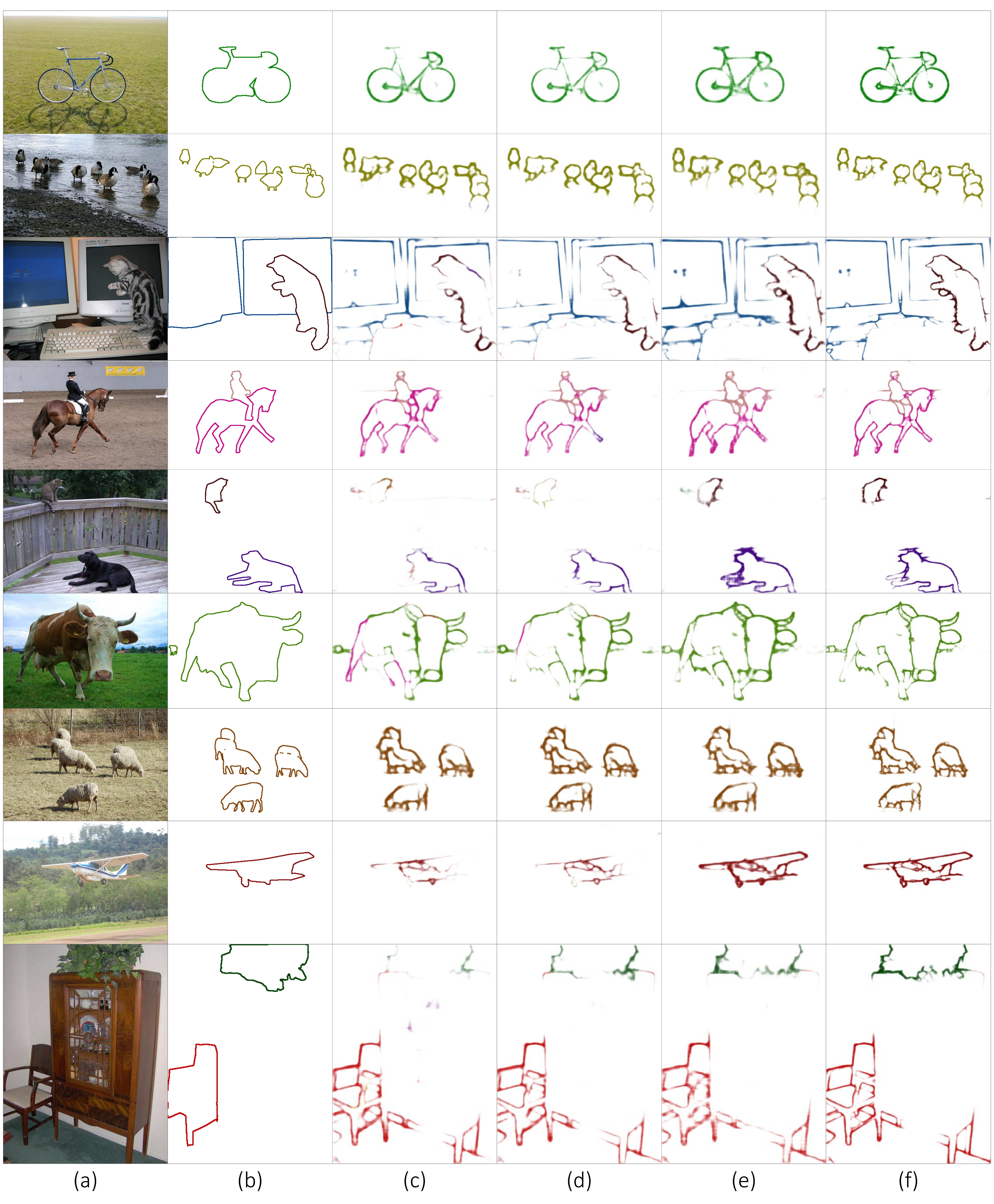}
\caption{
Qualitative results on the SBD benchmark \emph{test} set.
(a) Input image.
(b) Ground-truth semantic boundary label.
(c) Results of our final model using VGG16 backbone. 
(d) Results of our final model using VGG19 backbone. 
(e) Results of our final model using ResNet50 backbone. 
(f) Results of our final model using ResNet101 backbone. 
} 
\label{fig:final_qual}
\end{figure*}

\begin{figure*}
\begin{center}
\includegraphics[trim={3.4cm 6cm 4.2cm 6cm},clip, width=0.243 \linewidth] {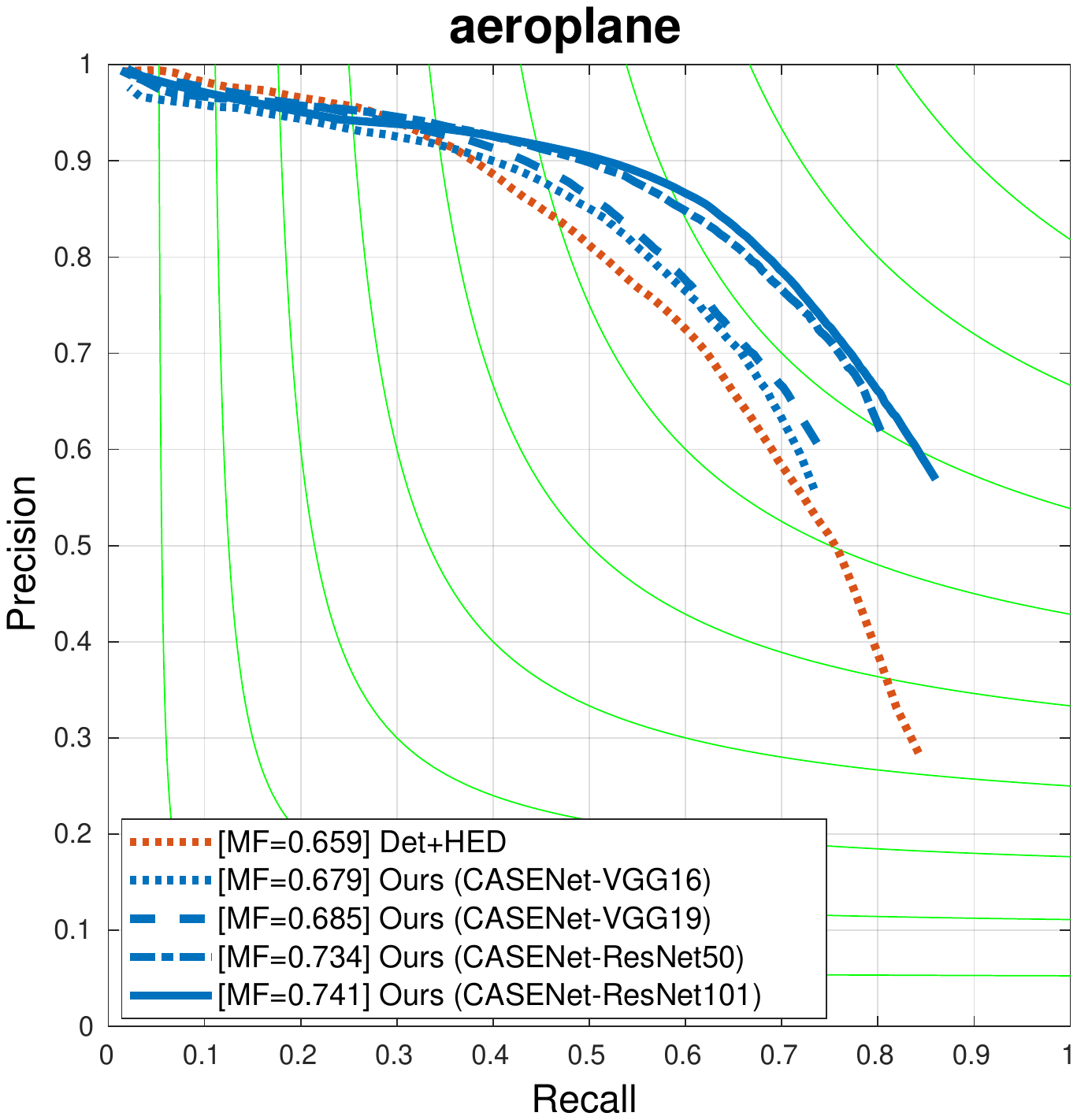}
\includegraphics[trim={3.4cm 6cm 4.2cm 6cm},clip, width=0.243 \linewidth] {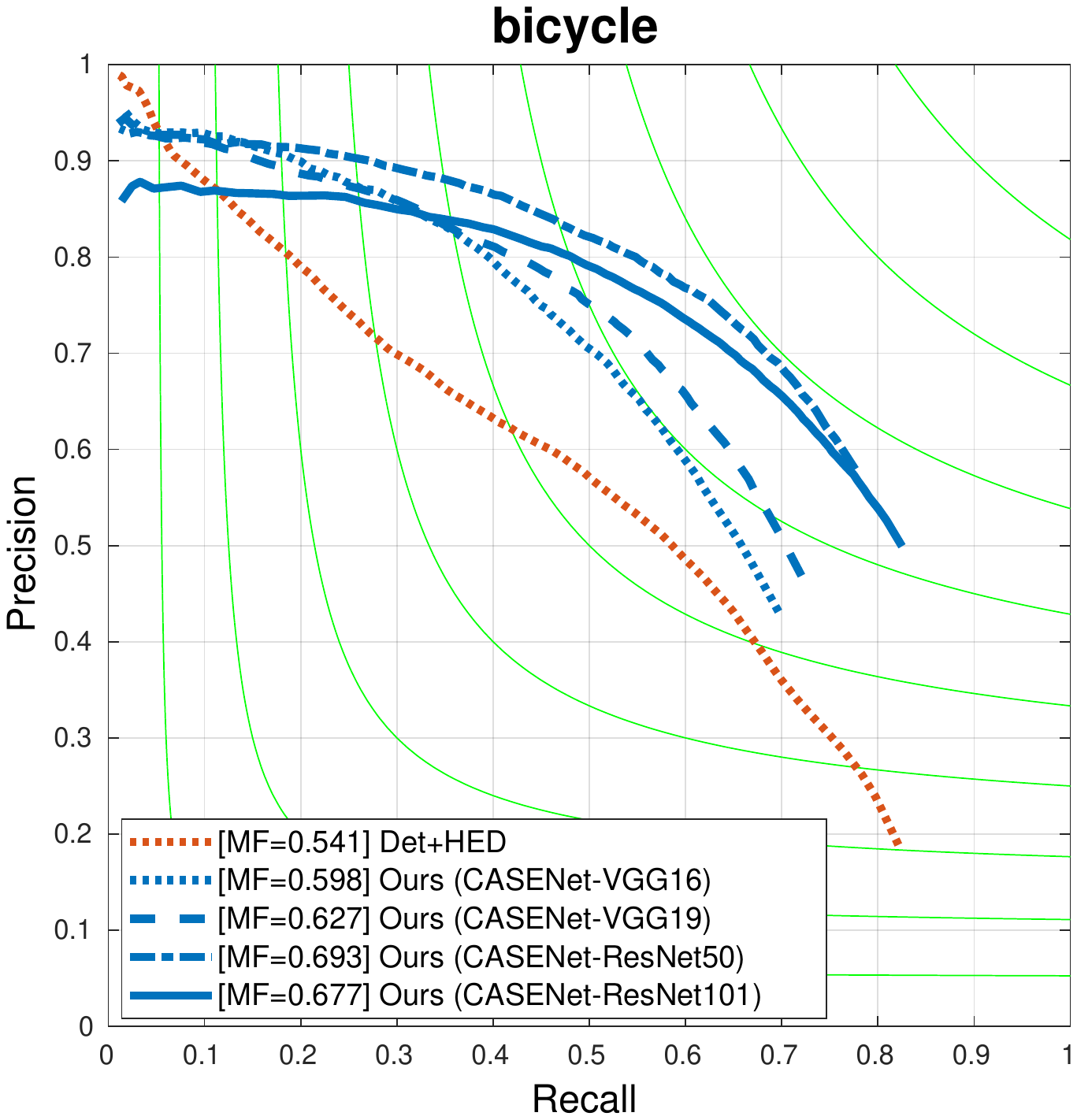}
\includegraphics[trim={3.4cm 6cm 4.2cm 6cm},clip, width=0.243 \linewidth] {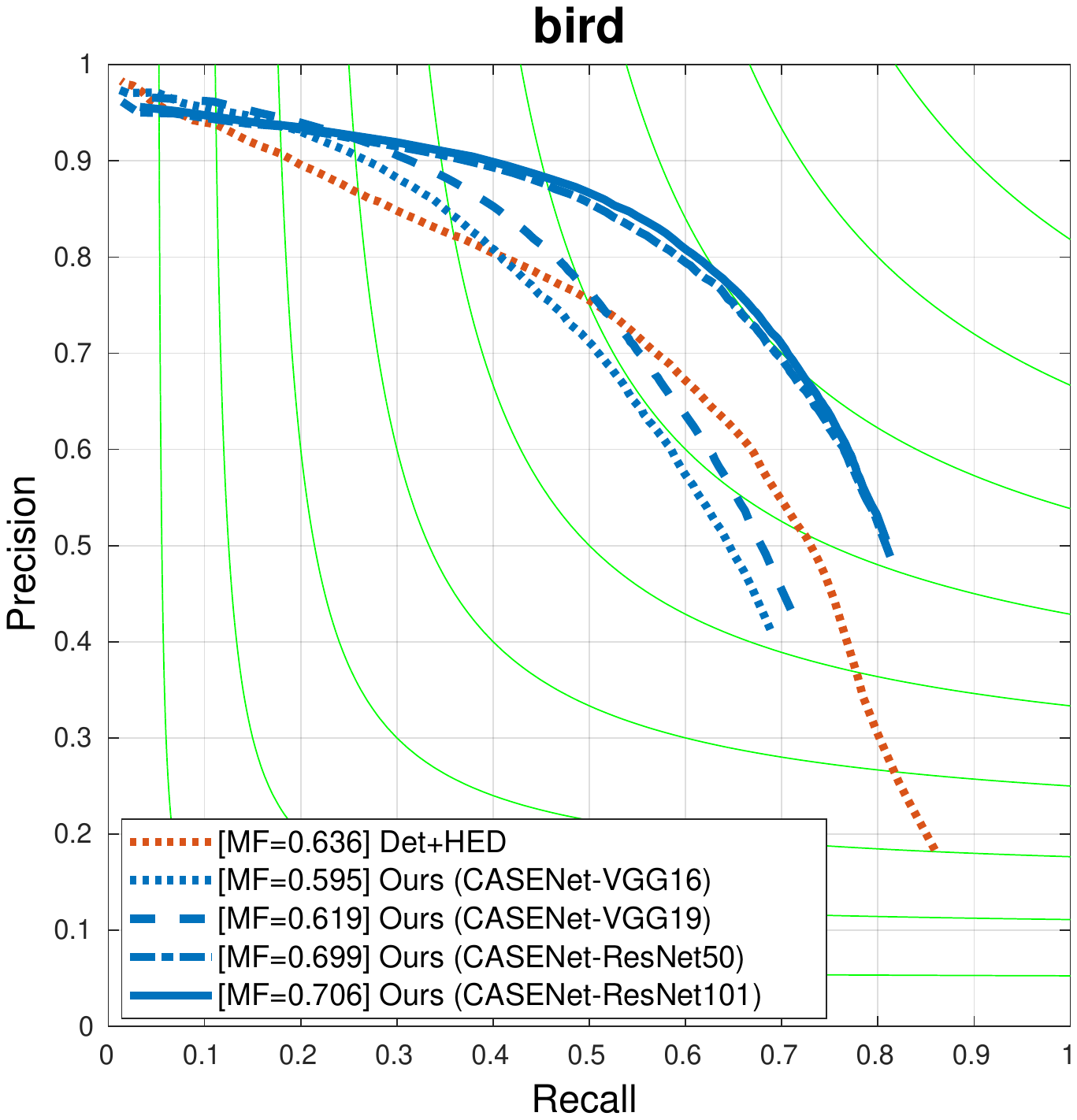}
\includegraphics[trim={3.4cm 6cm 4.2cm 6cm},clip, width=0.243 \linewidth] {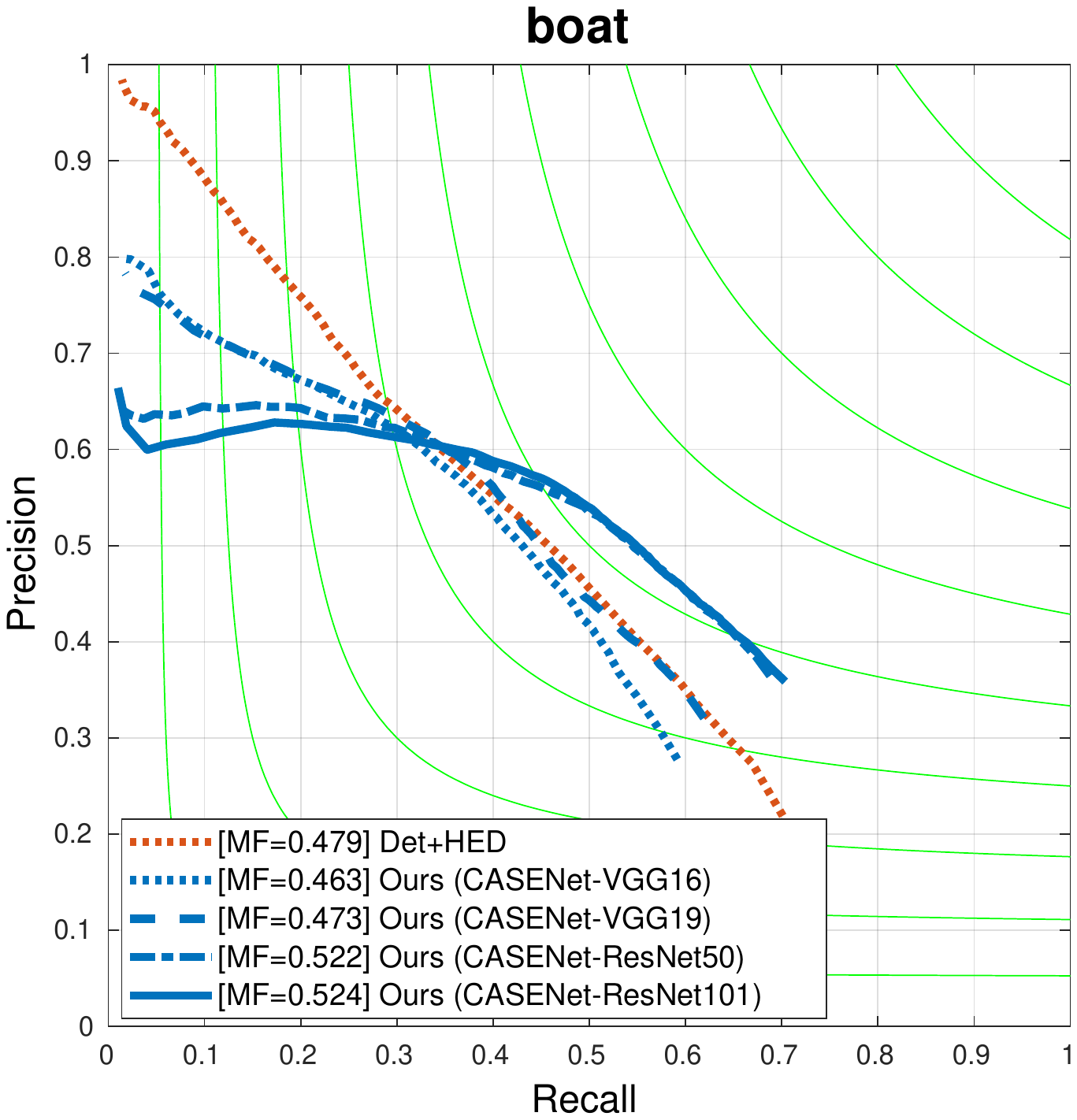}\\
\includegraphics[trim={3.4cm 6cm 4.2cm 6cm},clip, width=0.243 \linewidth] {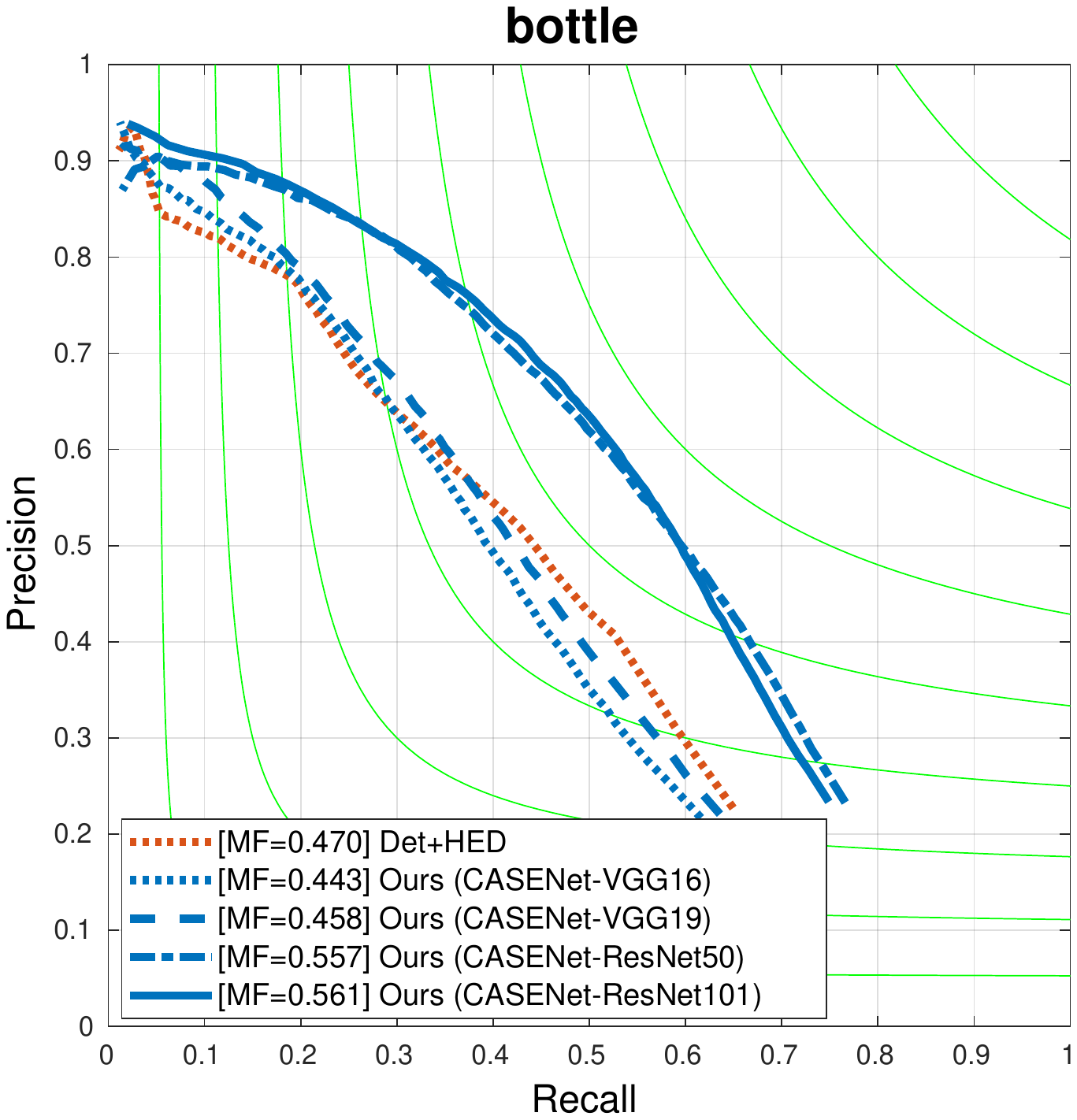}
\includegraphics[trim={3.4cm 6cm 4.2cm 6cm},clip, width=0.243 \linewidth] {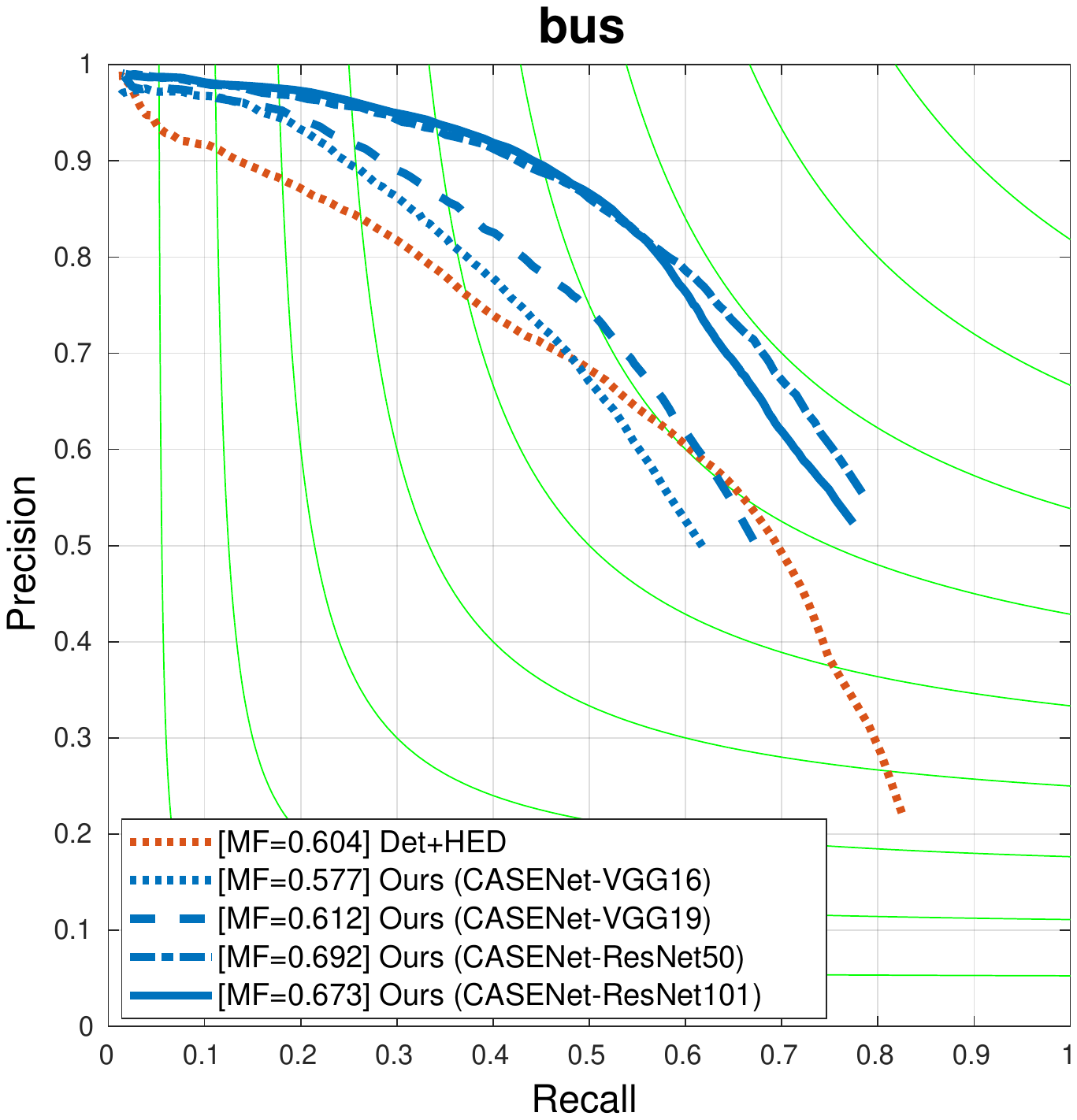}
\includegraphics[trim={3.4cm 6cm 4.2cm 6cm},clip, width=0.243 \linewidth] {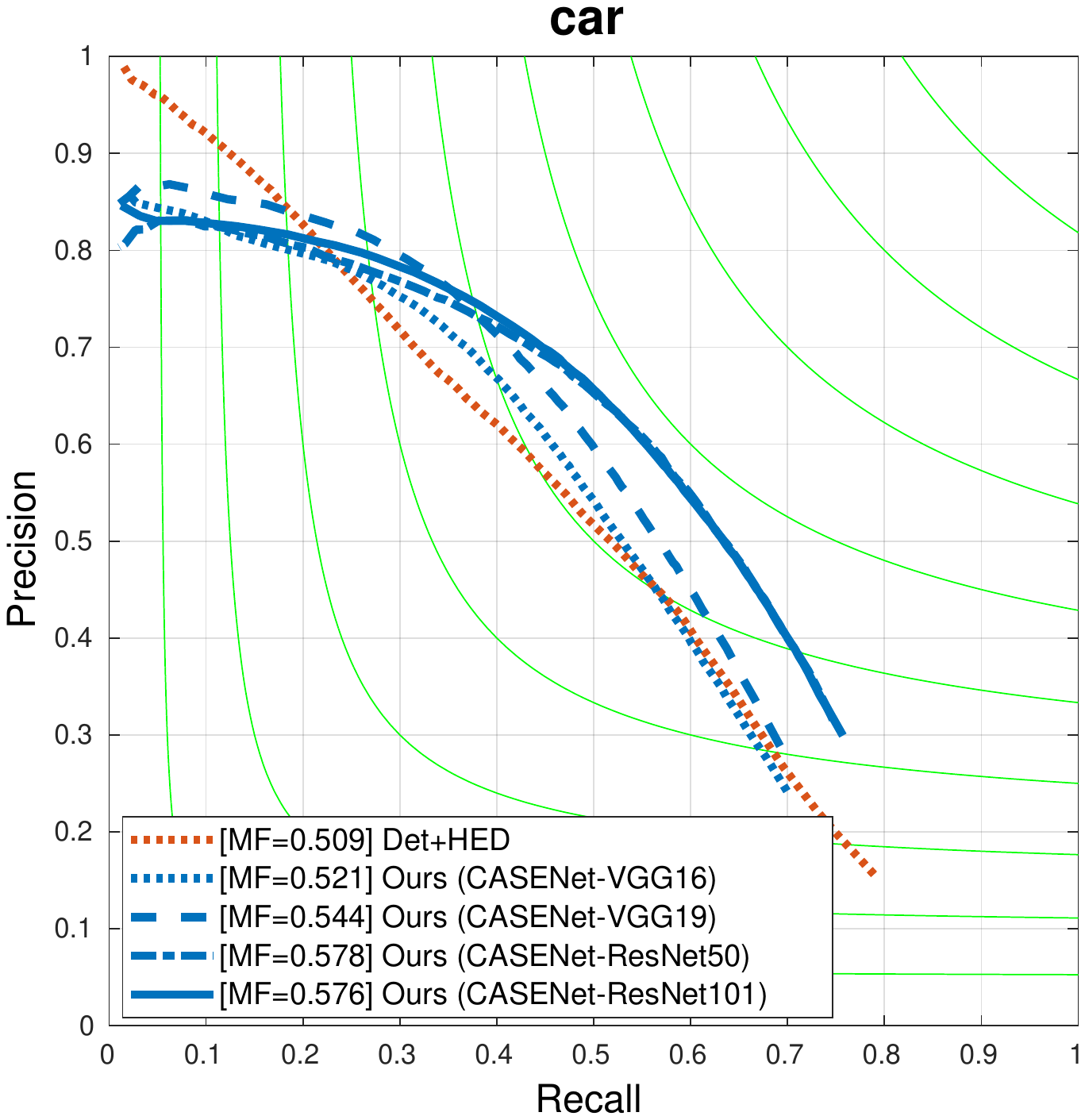}
\includegraphics[trim={3.4cm 6cm 4.2cm 6cm},clip, width=0.243 \linewidth] {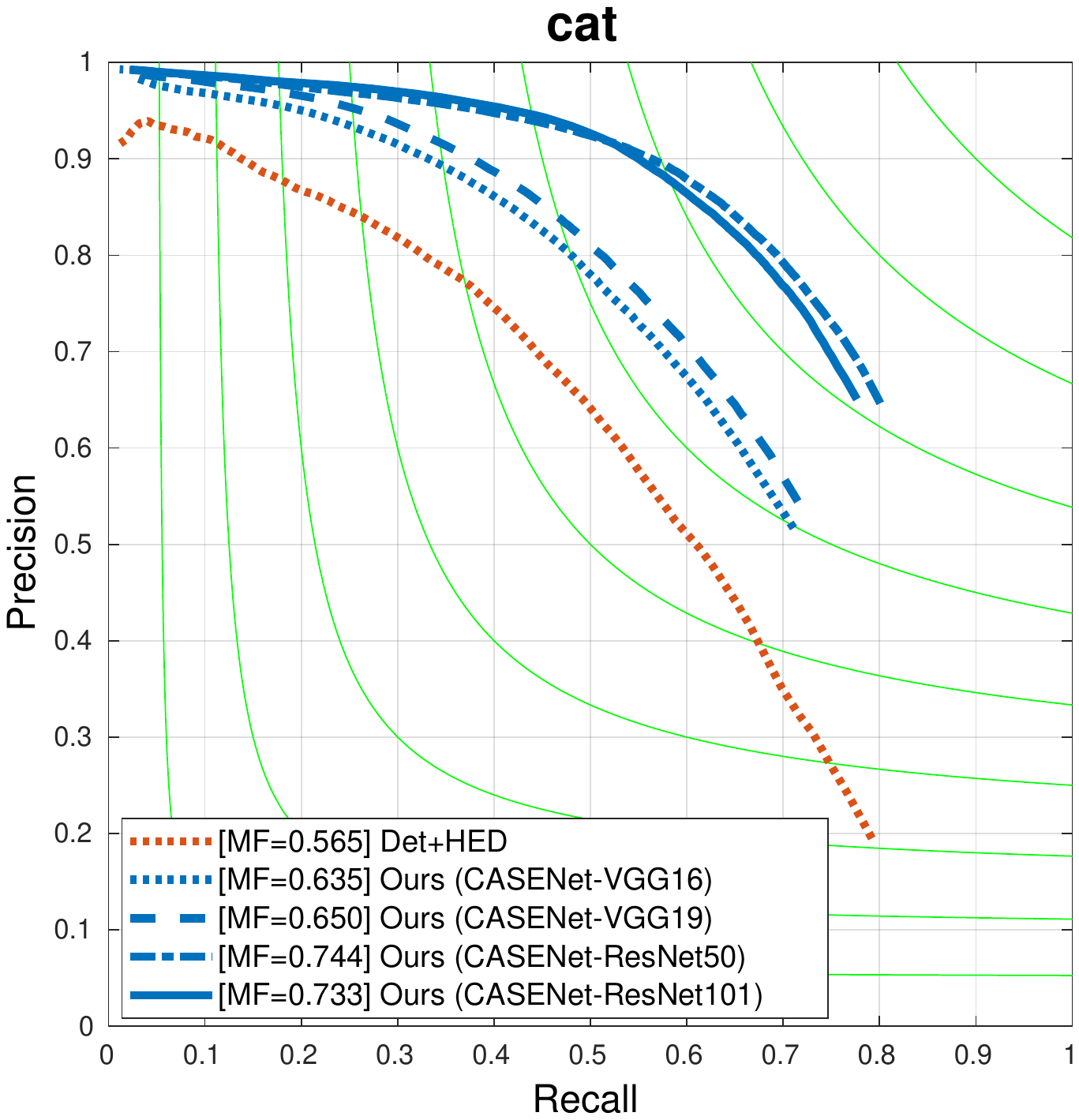}\\
\includegraphics[trim={3.4cm 6cm 4.2cm 6cm},clip, width=0.243 \linewidth] {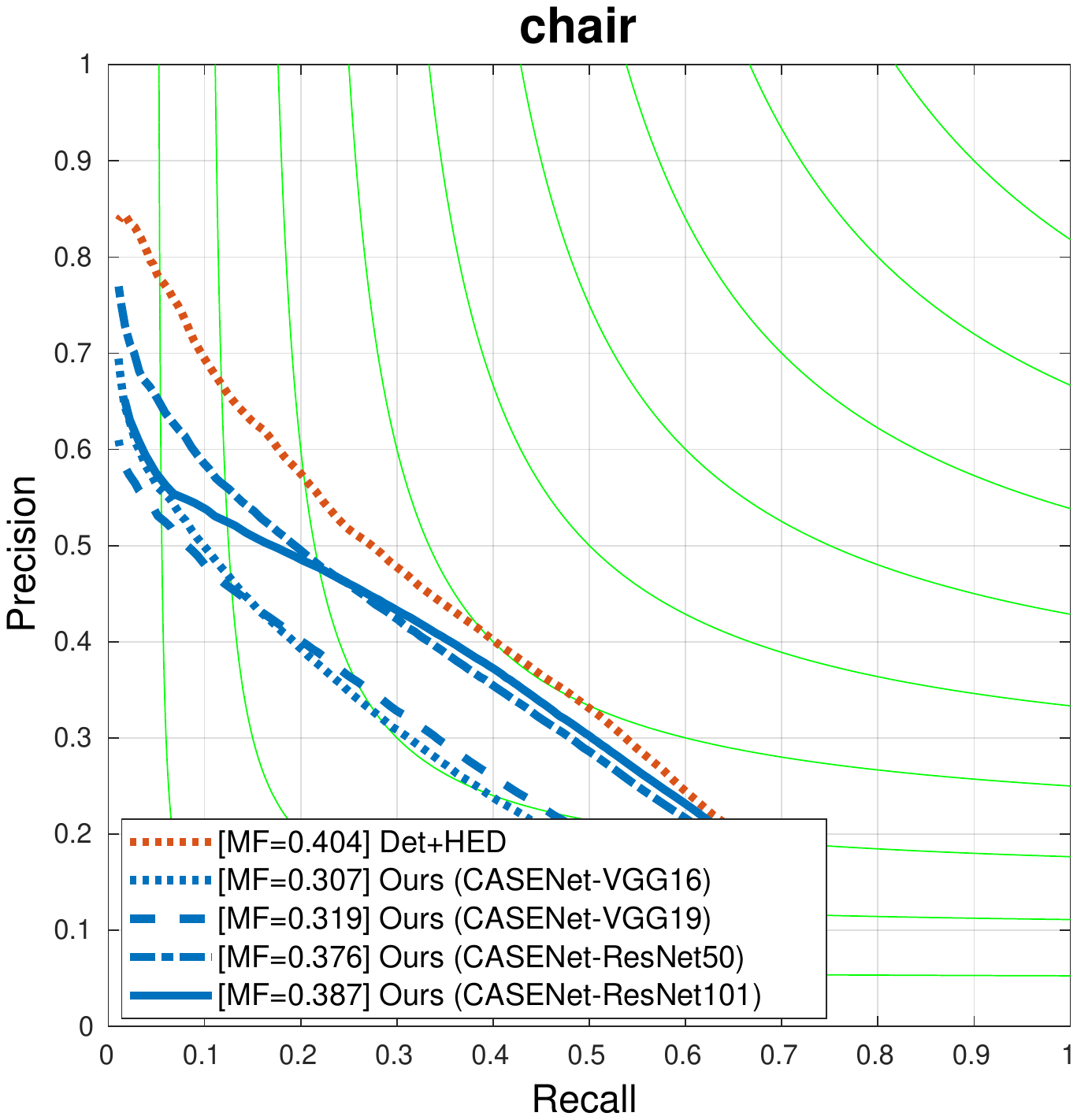}
\includegraphics[trim={3.4cm 6cm 4.2cm 6cm},clip, width=0.243 \linewidth] {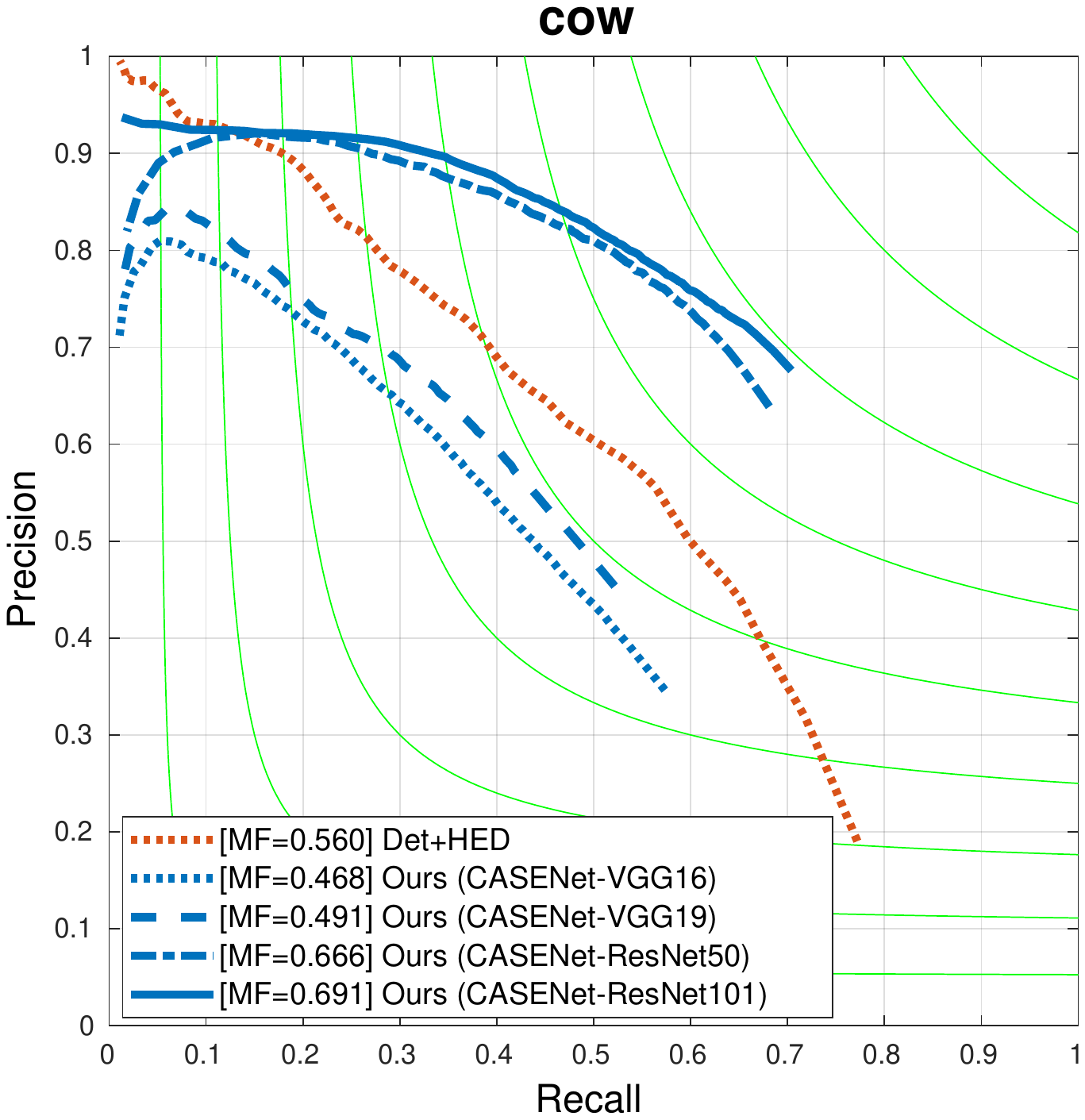}
\includegraphics[trim={3.4cm 6cm 4.2cm 6cm},clip, width=0.243 \linewidth] {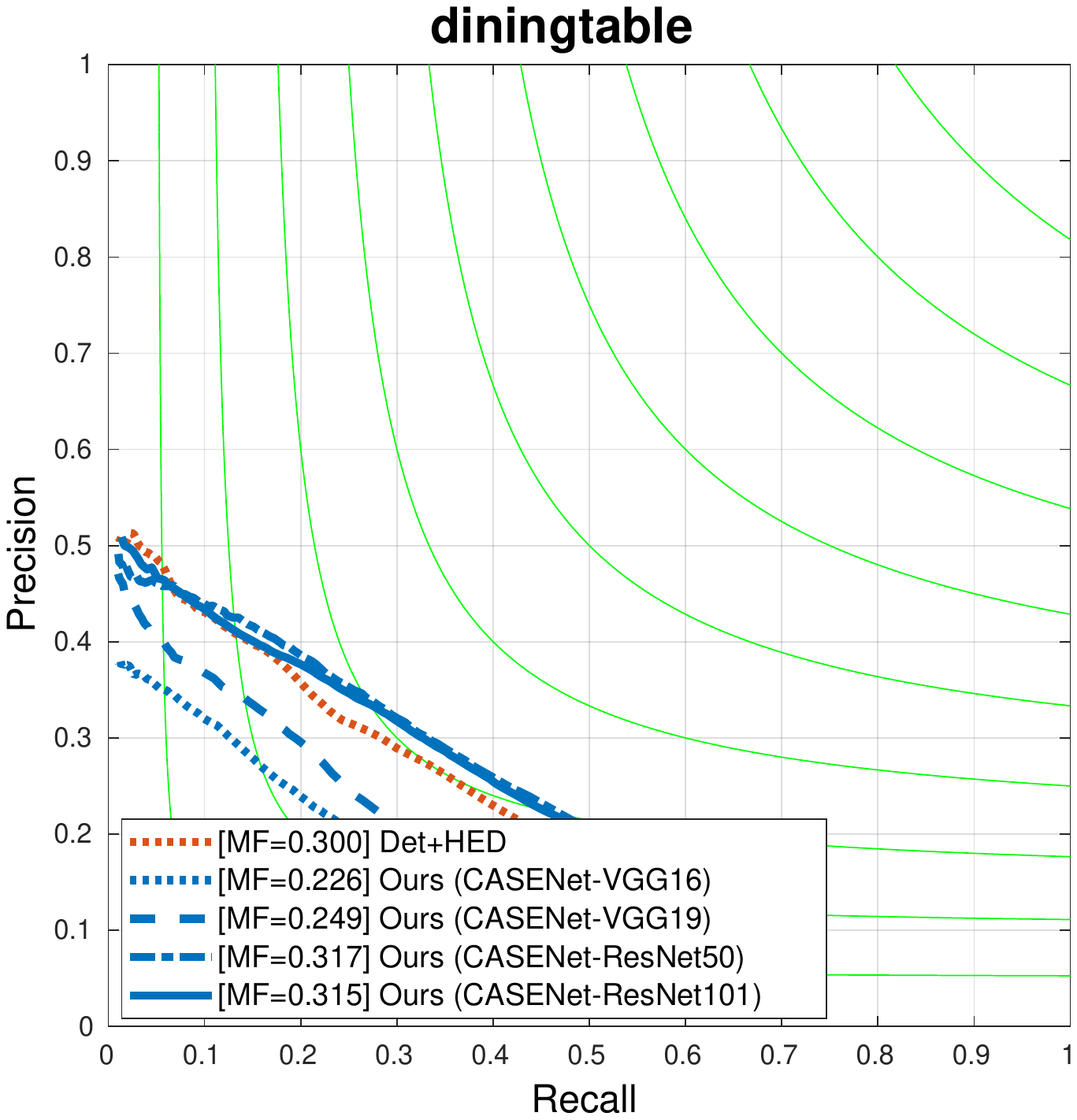}
\includegraphics[trim={3.4cm 6cm 4.2cm 6cm},clip, width=0.243 \linewidth] {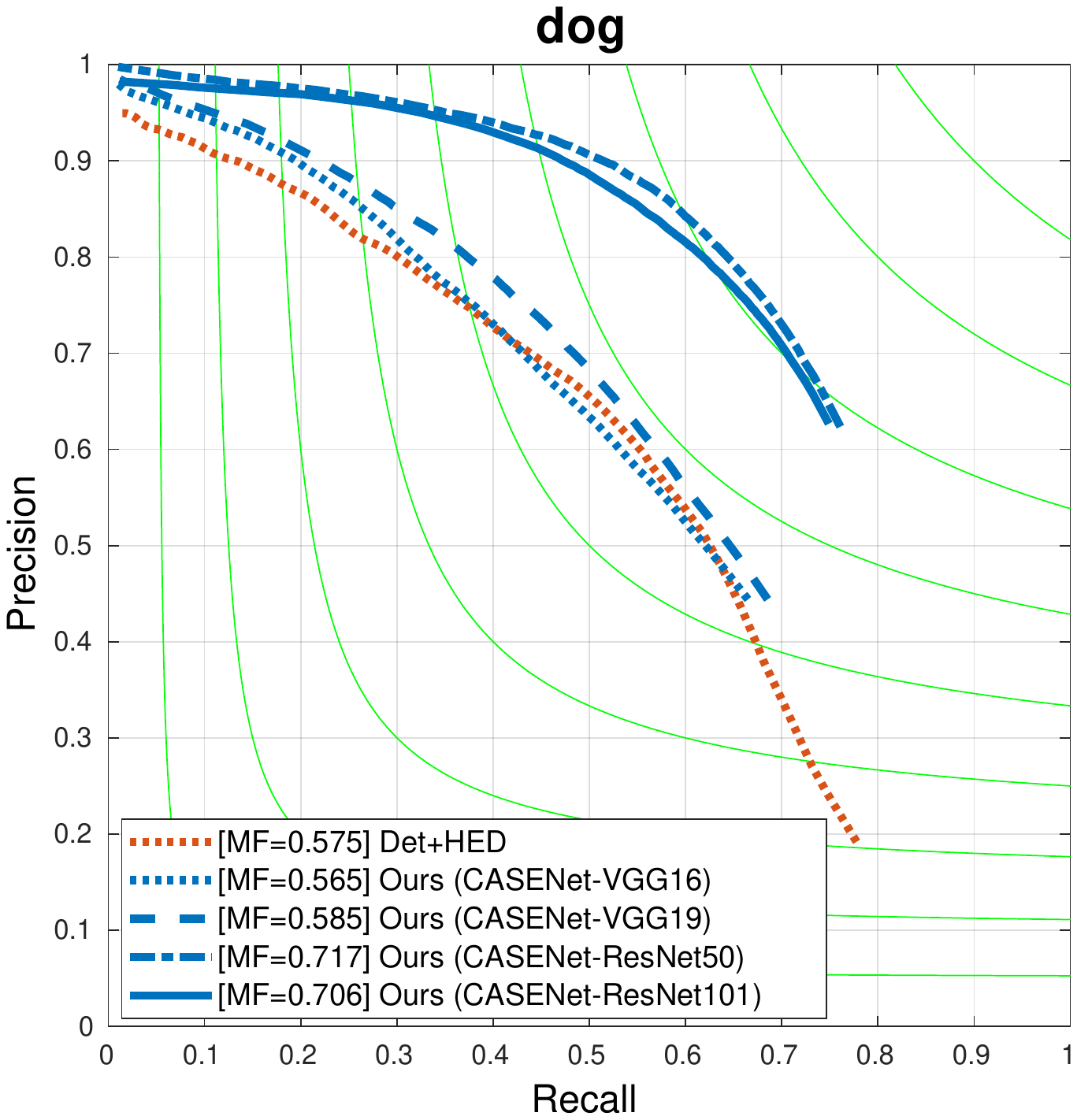}\\
\includegraphics[trim={3.4cm 6cm 4.2cm 6cm},clip, width=0.243 \linewidth] {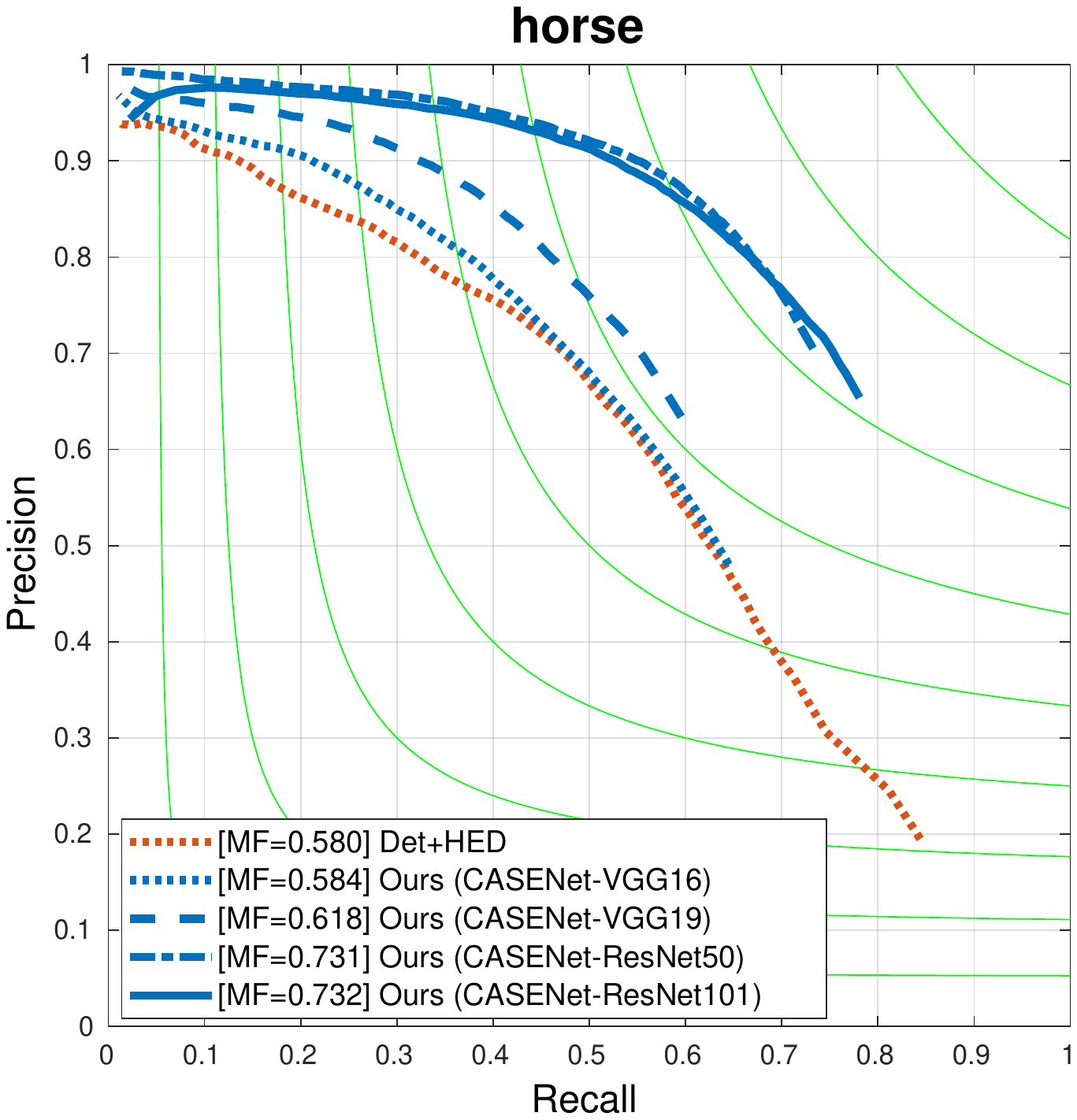}
\includegraphics[trim={3.4cm 6cm 4.2cm 6cm},clip, width=0.243 \linewidth] {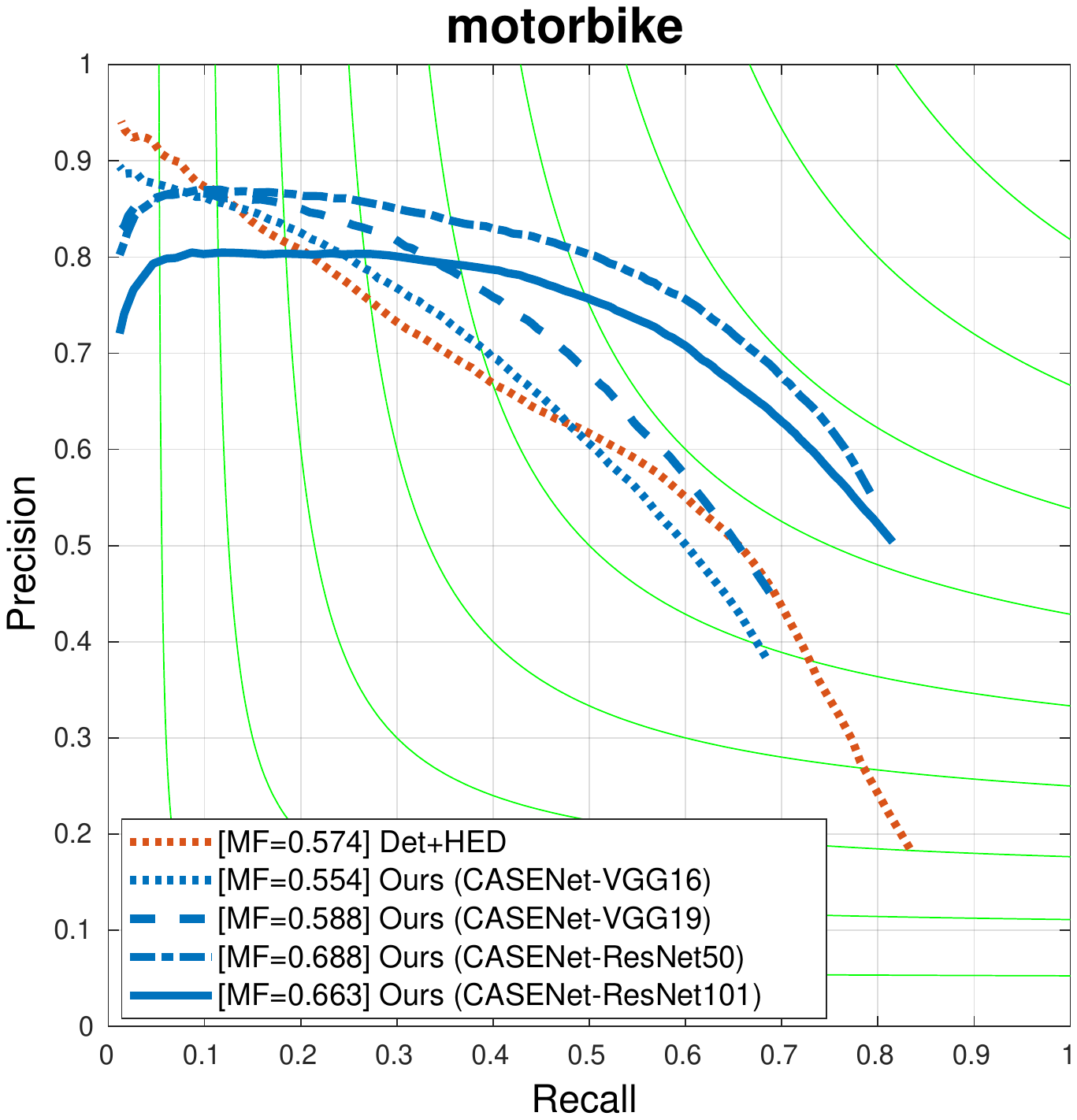}
\includegraphics[trim={3.4cm 6cm 4.2cm 6cm},clip, width=0.243 \linewidth] {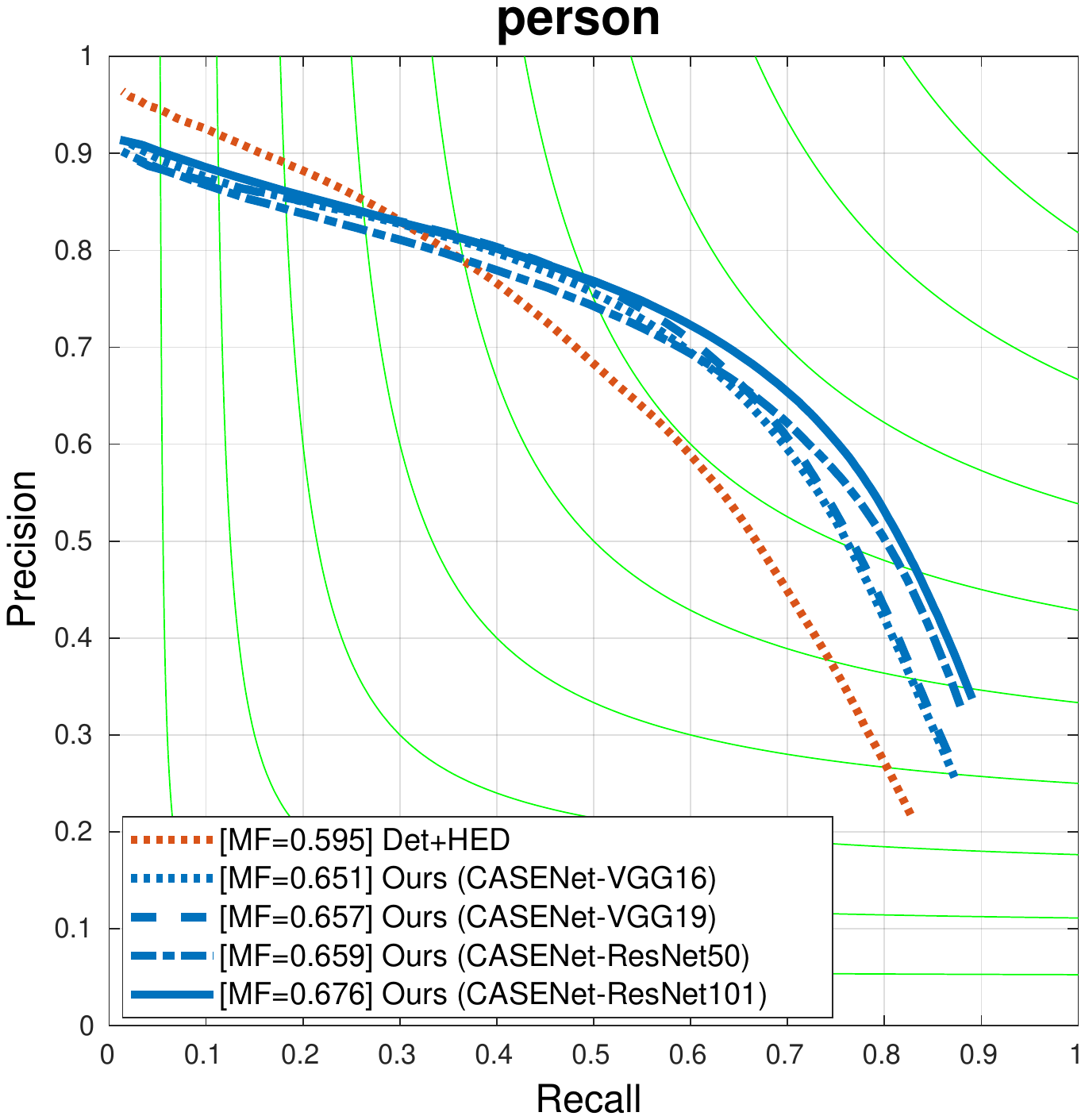}
\includegraphics[trim={3.4cm 6cm 4.2cm 6cm},clip, width=0.243 \linewidth] {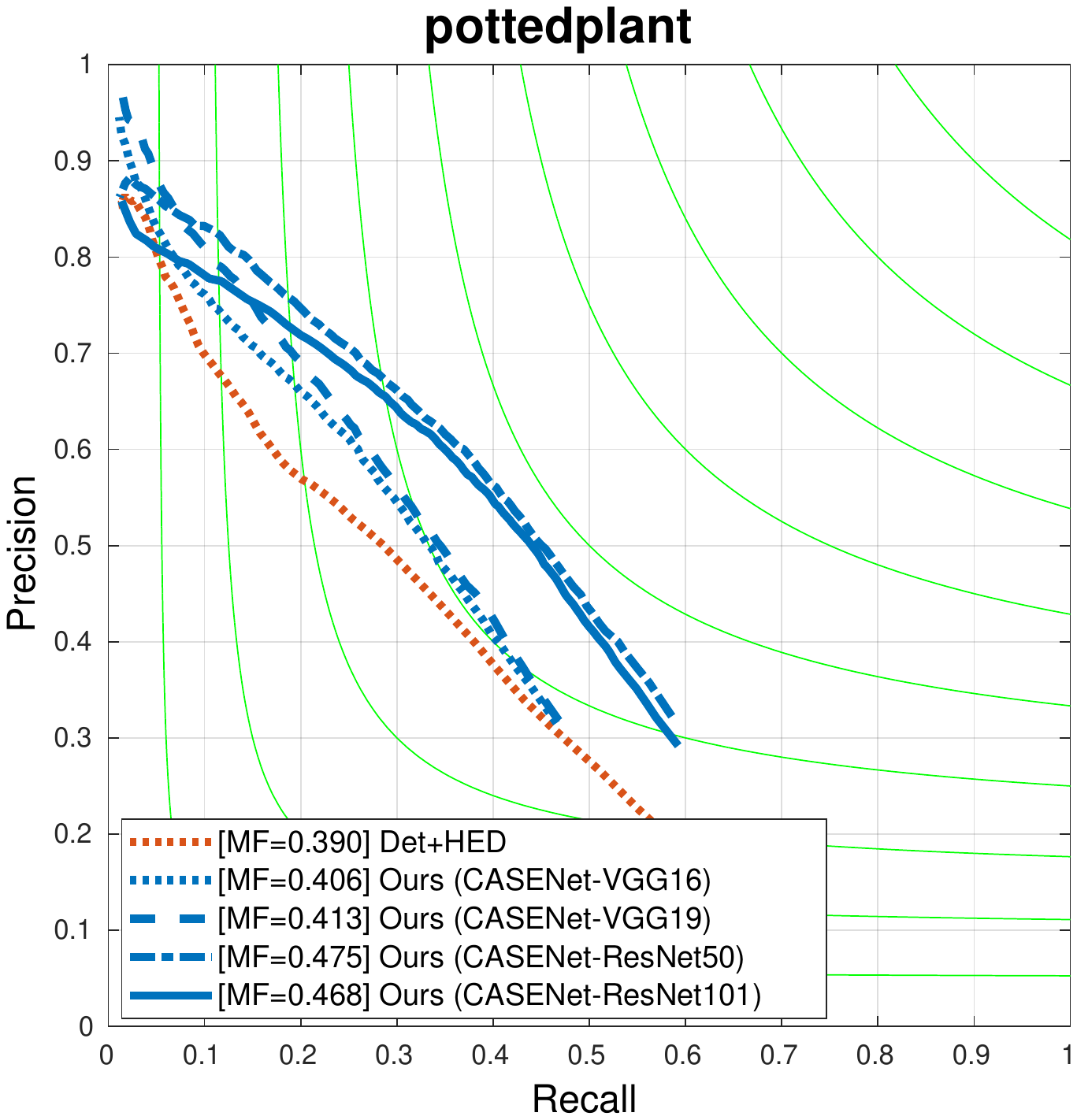}\\
\includegraphics[trim={3.4cm 6cm 4.2cm 6cm},clip, width=0.243 \linewidth] {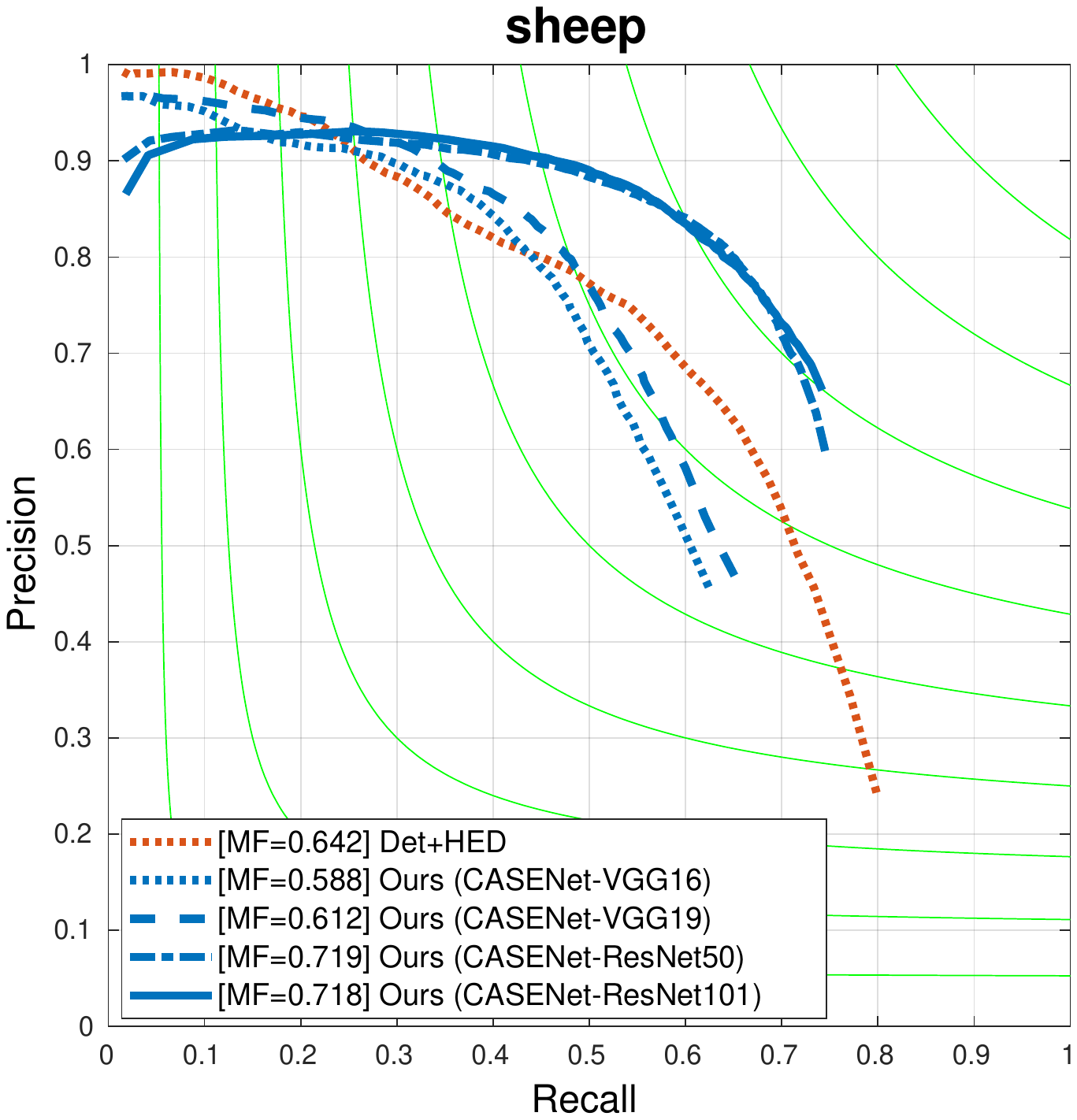}
\includegraphics[trim={3.4cm 6cm 4.2cm 6cm},clip, width=0.243 \linewidth] {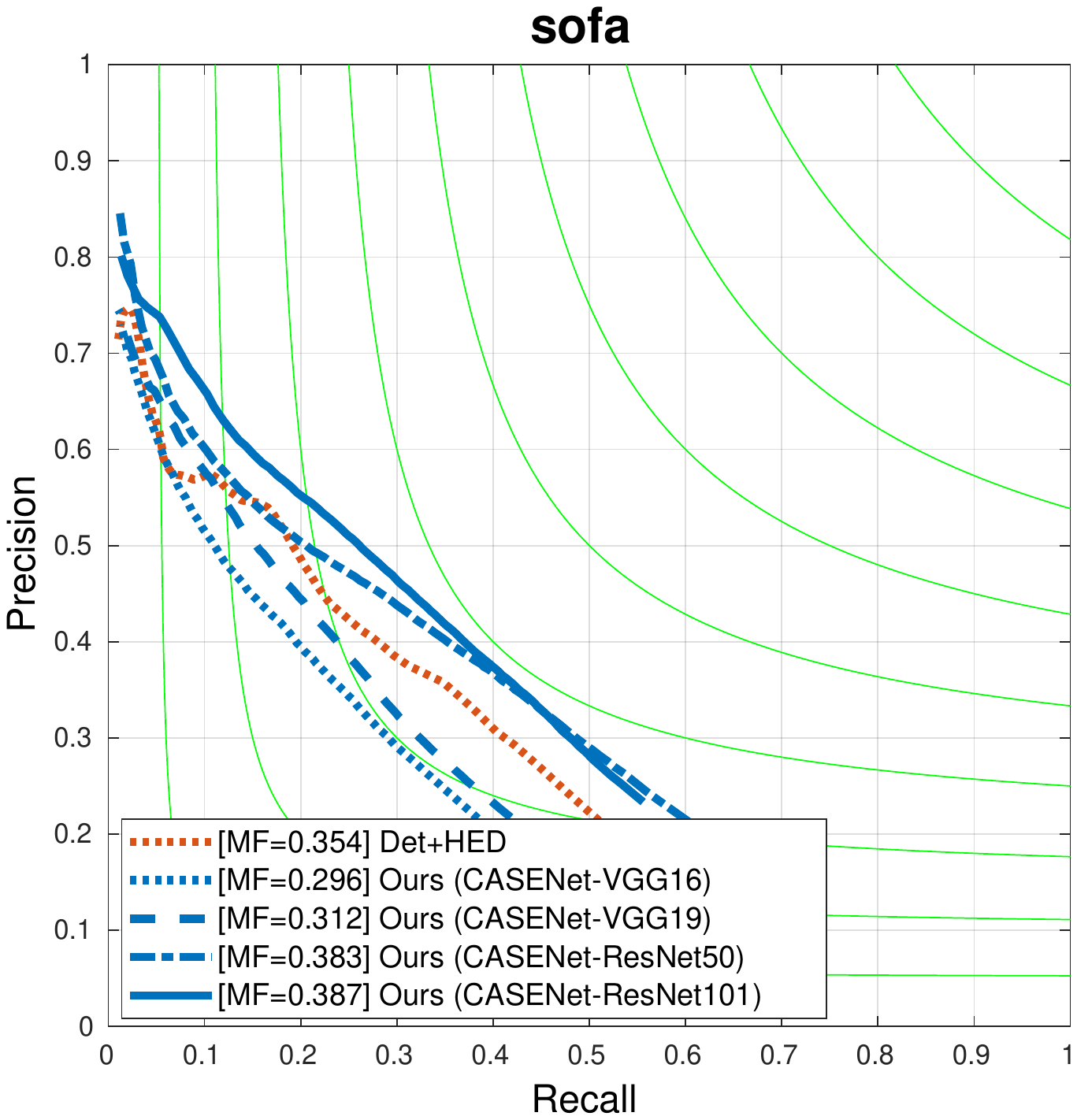}
\includegraphics[trim={3.4cm 6cm 4.2cm 6cm},clip, width=0.243 \linewidth] {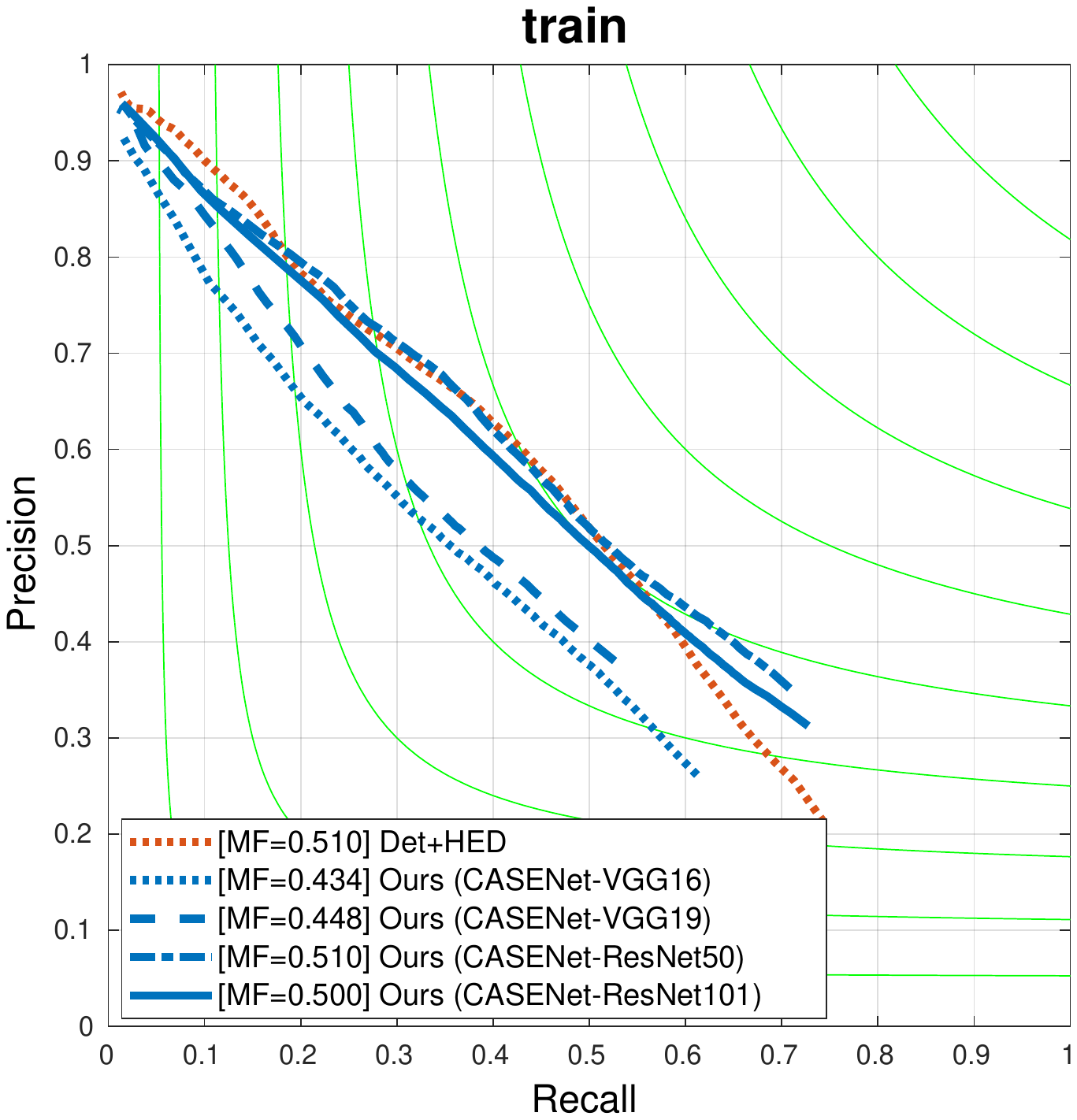}
\includegraphics[trim={3.4cm 6cm 4.2cm 6cm},clip, width=0.243 \linewidth] {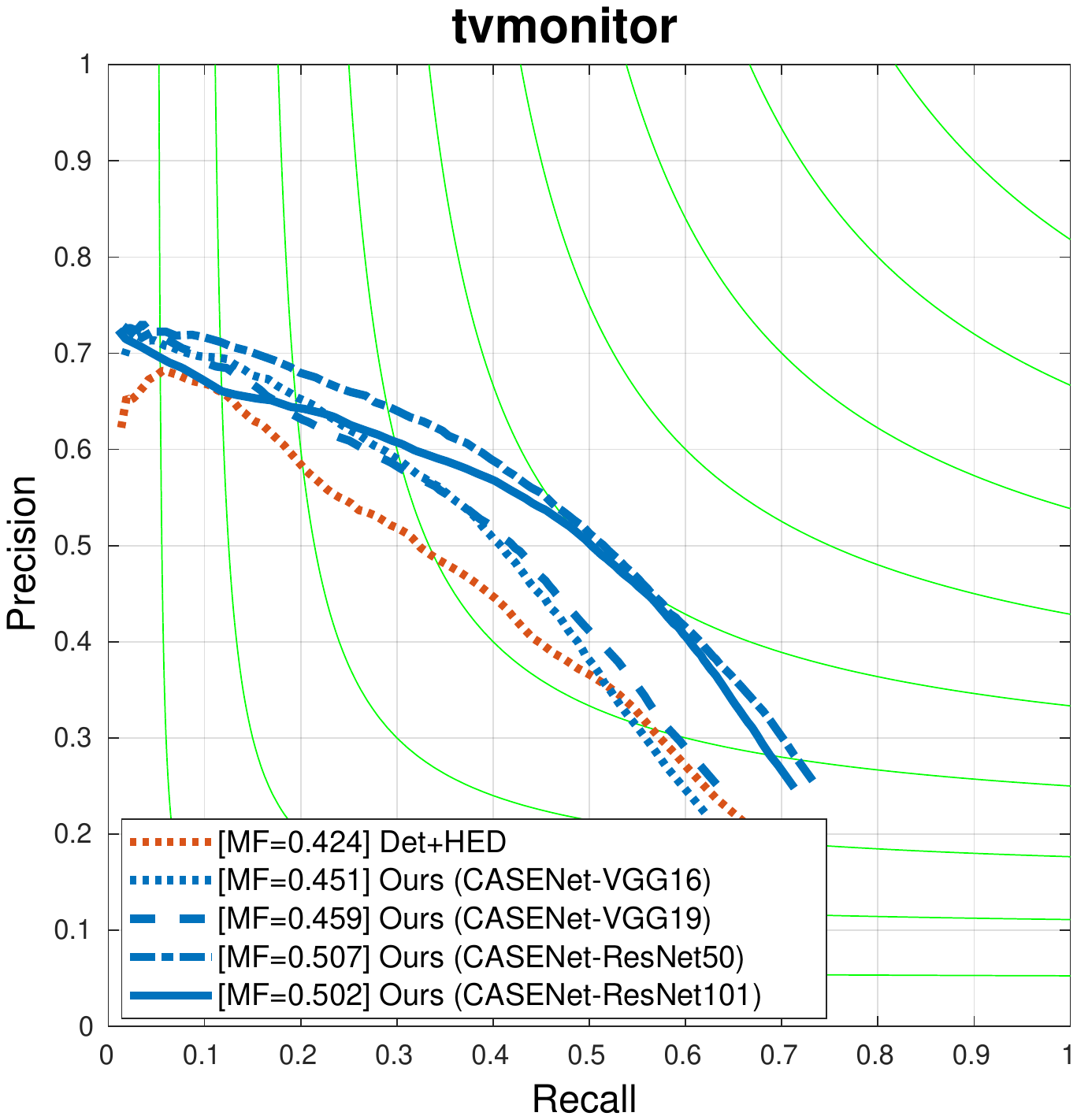}
\end{center}
\vspace{-3mm}
\caption{
PR curves of our final models using different backbones and Det+HED~\citep{Khoreva_2016_CVPR}.
MF scores of the methods are given in the legend.
} 
\label{fig:plot}
\end{figure*}

\subsection{Performance Analysis of Our Final Model}

\noindent{\textbf{Comparison to SBD models.}}
CASENet trained with our pseudo labels is denoted by Ours+CASENet and considered as our final model.
The performance of this model is evaluated and compared with previous records on the SBD benchmark~\citep{SBD} \emph{test} set.
As summarized in~\Tbl{sbd_test}, our final models achieve outstanding performance.
For example, our model based on the VGG16 backbone is as competitive as Det+HED \citep{Khoreva_2016_CVPR}, state-of-the-art weakly supervised method using the same backbone yet relying on stronger supervision,
and our model based on the VGG19 backbone catches up with 89\% of the performance of HFL-FC8~\citep{Bertasius_ICCV2015}, a fully supervised model using the same backbone network.
The performance of the fully supervised CASENet can be regarded as the upper-bound we can achieve by weakly supervised learning, and our method recovers 88\% and 84\% of the upper-bound with VGG and ResNet backbones, respectively.

We also train DFF~\citep{dff} with our pseudo labels to demonstrate the universality of our method. 
The trained model, denoted by Ours+DFF, achieves 1\% improvement in MF compared to Ours+CASENet.
This result shows that our pseudo labels can be used to train a variety of SBD models, and that the performance of our final model increases by incorporating stronger SBD models.

\Fig{final_qual} shows qualitative results of our final model with different backbone networks.
Although the proposed framework demands image-level class labels only as supervision, our final model exhibits promising results even for images with multiple instances of various classes and background clutters. 
Also, a stronger backbone can help estimate crisp boundaries, classify their labels correctly, and suppress internal edges of objects. 
In addition, we provide PR curves of our final models and
Det+HED~\citep{Khoreva_2016_CVPR} in \Fig{plot} 
for a more thorough comparison.
In the figure, green lines indicate the same value of F-measure.
Overall, our models based on ResNet backbones outperform Det+HED, although they demand weaker level of supervision.
Also, our model based on VGG16 backbone is as competitive as Det+HED, and even better for some categories like {\it bicycle}, {\it cat}, {\it tvmonitor}, and {\it person}.

\begin{figure}[!t]
\includegraphics[width = 0.98\linewidth]{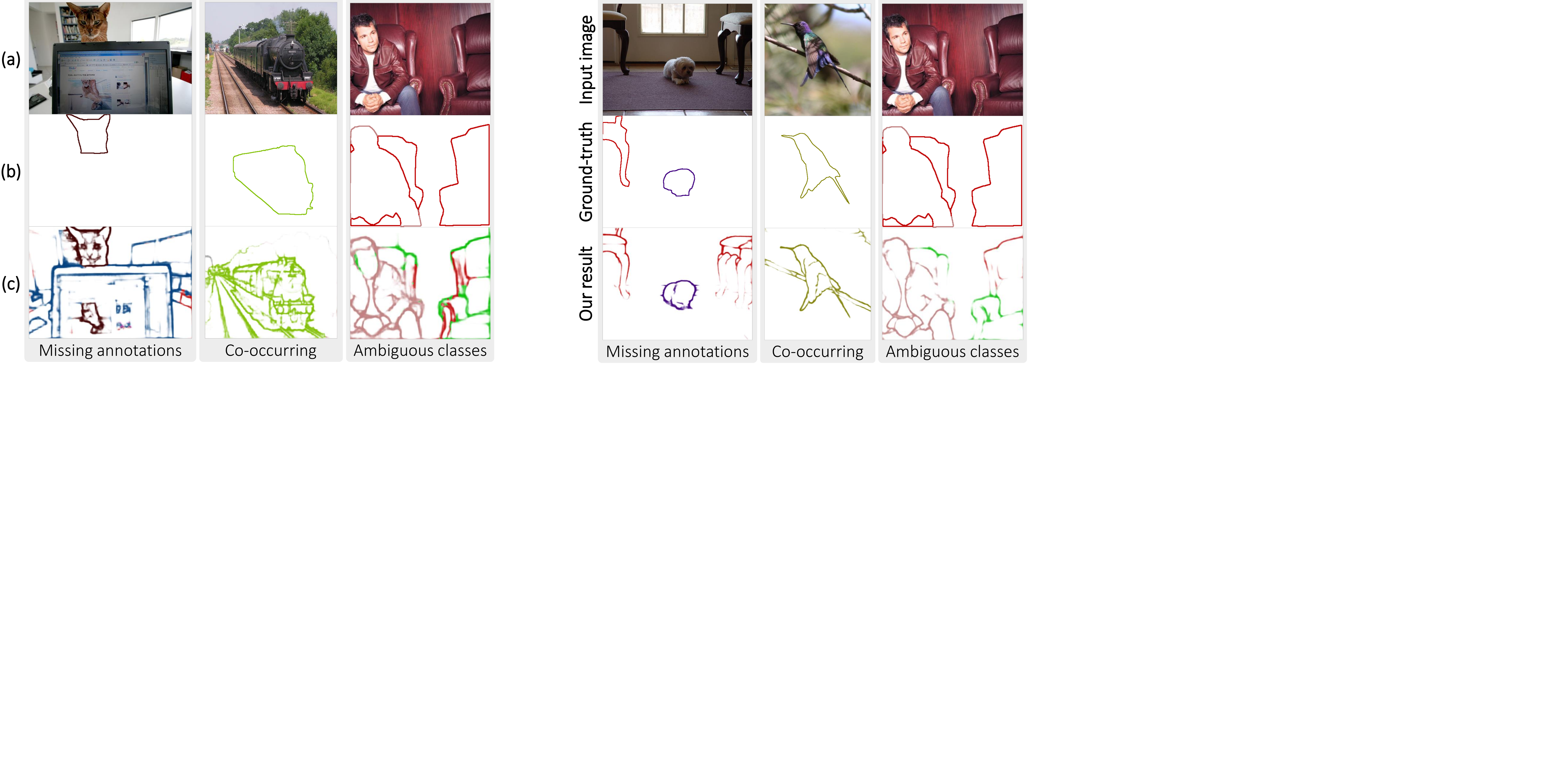}
\vspace{-2mm}
\caption{Failure cases on the SBD benchmark \emph{test} set.
} 
\label{fig:final_qual_fail}
\end{figure}

\noindent{\textbf{Evaluation on the Cityscapes dataset.}}
The efficacy of our method is also demonstrated on the Cityscapes dataset.
Unlike the SBD dataset, the Cityscapes dataset could contain a large number of diverse object classes in each image, and defines background components that appear in almost all images (\eg, \emph{road}, \emph{sky}, \emph{vegetation}) as classes to be recognized.
Hence, on the Cityscapes dataset, multi-class image classification is tricky and CAM does not work as desired accordingly, meaning that 
it is not straightforward to learn WSBDN directly on the Cityscapes dataset.
For the same reason, to the best of our knowledge, no weakly supervised learning method using image-level class labels has been applied to the Cityscapes dataset.

\begin{figure}[!t]
\begin{center}
\includegraphics[width=1 \linewidth]{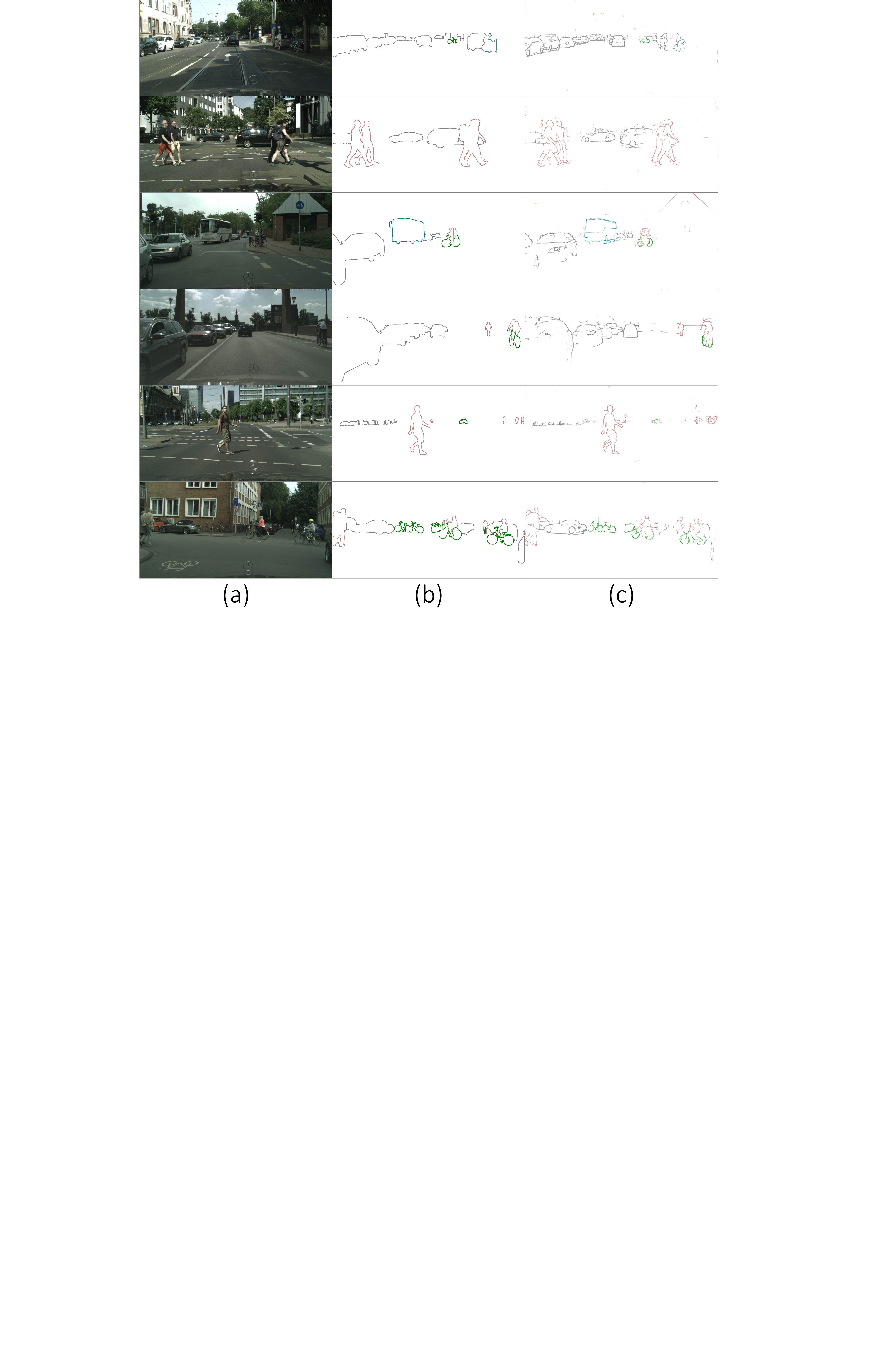}
\end{center}
\vspace{-5mm}
\caption{Qualitative results of the 6 foreground classes on the Cityscapes dataset \emph{val} set.
(a) Input image.
(b) Ground-truth semantic boundary label.
(c) Results of Ours+CASENet trained on the SBD dataset. 
}
\label{fig:qual_city}
\vspace{-3mm}
\end{figure}

Therefore, we instead apply our final models trained on the SBD dataset to the Cityscapes dataset, without finetuning, and evaluate its boundary detection accuracy for 6 foreground classes (\ie, \emph{bike, bus, car, motorbike, person, train}) the two datasets have in common.
As shown in Tab.~\ref{tab:eval_cityscape}, our models catch up with 82.7\% and 87.5\% of their fully supervised counterpart trained on the SBD dataset.
Our method performs well in this domain transfer setting, except for the \emph{train} class whose appearance is substantially different between the two datasets.
\Fig{qual_city} shows qualitative results of Ours+CASENet on the Cityscapes dataset.
Although it is transferred from another dataset, our model shows promising results even for images with complex road scenes containing multiple object classes with occlusions.

\begin{table} \small
\caption{The quality of predicted semantic boundaries of the 6 foreground classes on the Cityscapes \emph{val} set in MF (\%). The type of supervision (S.) indicates $\mathcal{F}_{s}$--pixel-level semantic boundary of SBD dataset and $\mathcal{I}_{s}$--image-level class label of SBD dataset.
}
\centering
\scalebox{0.93}{
\begin{tabular}{@{}C{2.4cm}@{}@{}C{0.6cm}@{}|@{}C{0.75cm}@{}@{}C{0.75cm}@{}@{}C{0.75cm}@{}@{}C{0.75cm}@{}@{}C{1.1cm}@{}@{}C{0.8cm}@{}|@{}C{1cm}@{}}
    \toprule
    Method\raggedright        & S. & bike & bus & car & mbk & person & train & mean\\
    \midrule
    CASENet\raggedright       & $\mathcal{F}_{s}$ & 70.2 & 50.0 & 85.9 & 47.6 & 82.2 & 15.2 & 58.5 \\
    DFF\raggedright           & $\mathcal{F}_{s}$ & 69.2 & 47.5 & 86.0 & 42.1 & 84.6 & 10.0 & 56.6 \\
    \midrule
    Ours+CASENet\raggedright  & $\mathcal{I}_{s}$ & 59.5 & 40.0 & 72.5 & 41.0 & 72.5 & 4.8 & 48.4 \\
    Ours+DFF\raggedright      & $\mathcal{I}_{s}$ & 61.9 & 40.7 & 73.0 & 43.2 & 74.7 & 3.8 & 49.5 \\
    \bottomrule
\end{tabular}}
\label{tab:eval_cityscape}
\vspace{-2mm}
\end{table}

\vspace{1mm}
\noindent{\textbf{Failure cases.}}
\Fig{final_qual_fail} presents a few failure examples of our final model.
As can be seen in the left example of Fig.~\ref{fig:final_qual_fail}, some failure cases are caused by incomplete labels missing true object boundaries.
Our model often falsely detects some patterns that are irrelevant to the predefined object classes but frequently co-occur with them like rails under trains and tree branches around birds in the middle column of the figure.
Also, ambiguity between classes, like sofa and chair in the right column, could cause confusion in prediction.

\vspace{1mm}
\noindent{\textbf{Comparison to weakly supervised semantic segmentation.}}
Since semantic segmentation and SBD are dual problems, we compare our model to state-of-the-art weakly supervised semantic segmentation methods \citep{affinitynet,IRNet,chen2020weakly,li2020group} relying on image-level supervision.
For fair comparisons, we adopt ResNet50 backbone for AffinityNet~\citep{affinitynet} instead of its original ResNet38~\citep{Wu2016WiderOD} backbone, and for the other methods use their original architectures as-is.
Following the protocol used in~\cite{Acuna_2019_CVPR}, their segmentation masks are converted to semantic boundaries through Sobel filtering.
Then, CASENets with the ResNet101 backbone are trained using their outputs as pseudo labels, as in our framework.

\begin{figure*} [!t]
\centering
\includegraphics[width = 0.96 \textwidth]{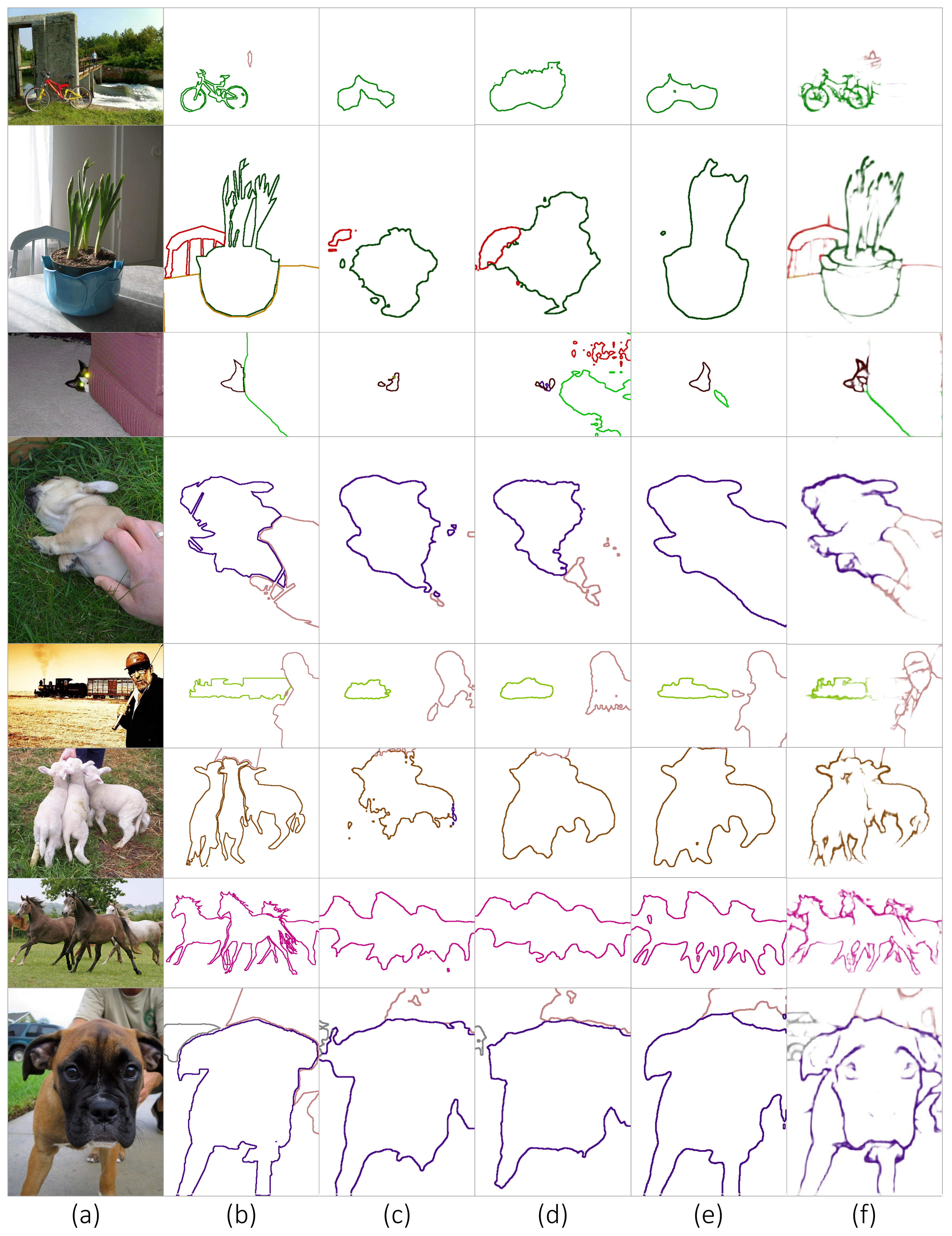}
\vspace{-2mm}
\caption{Qualitative comparison to weakly supervised semantic segmentation models on the PASCAL VOC 2012 \emph{val} set. (a) Input image. (b) Ground-truth semantic boundary label. (c) AffinityNet~\citep{affinitynet}. (d) IRNet~\citep{IRNet}. (e) GWSM~\citep{li2020group}. (f) Ours+CASENet. 
}
\label{fig:qual_pascal_voc}
\end{figure*}

The comparison is done on the PASCAL VOC 2012 dataset~\citep{Pascalvoc}.
As shown in~\Tbl{eval_pascal_voc}, our model outperforms all the semantic segmentation methods in the quality of semantic boundaries.
Note that our model outperforms~\cite{chen2020weakly} by a large margin although their model is trained with boundary information explicitly.
Qualitative results in \Fig{qual_pascal_voc} show that the segmentation methods have trouble when objects have complicated shapes (\eg, \emph{bicycle} and \emph{potted plant}) or are less discriminative (\eg, \emph{hand}) while our method detects boundaries of such objects effectively.

\begin{table} [!t] 
\caption{Semantic boundary detection and segmentation performance on the PASCAL VOC 2012 \emph{val} set in mean MF (\%) and mean Intersection over Union (\%), respectively.
}
\footnotesize
\centering
\begin{tabular}{
@{}C{3.4cm}@{}C{2.15cm}@{}C{2.25cm}@{}
}
    \toprule
    & Semantic Boundary & Semantic Segmentation \\
    \midrule
    \cite{affinitynet} & 46.5 & 56.8 \\
    \cite{IRNet}             & 51.1 & 63.9 \\
    \cite{chen2020weakly}     & 55.5 & 65.7 \\
    \cite{li2020group}             & 57.3 & 68.2 \\
    \midrule
    Ours                   & \textbf{60.0} & -- \\
    \bottomrule
\end{tabular}
\label{tab:eval_pascal_voc}
\vspace{-3mm}
\end{table}

\vspace{-3mm}
\section{Conclusion}
\label{sec:conclusion}

This paper has presented the first attempt to learn SBD using image-level class labels.
Even in this minimally supervised setting, semantic boundary labels can be reliably estimated for a group of pixels on a short line segment, and a MIL framework using the line segment labels can be used to train a SBD model effectively.
We also have introduced WSBDN, a new CNN architecture well suited to learn SBD reliably under the weak and uncertain supervision.
Our final model trained with pseudo labels generated by WSBDN has achieved impressive performance on a public benchmark for SBD.

\bibliographystyle{spbasic}      %
\bibliography{cvlab_kwak}

\end{document}